\documentclass[10pt,journal,compsoc]{IEEEtran}
\usepackage{color}
\usepackage{caption}
\usepackage{pbox}
\usepackage{multirow}

\newcommand{\RNum}[1]{\uppercase\expandafter{\romannumeral #1\relax}}

\ifCLASSOPTIONcompsoc
\usepackage[nocompress]{cite}
\else
\usepackage{cite}
\fi

\ifCLASSINFOpdf
\usepackage[pdftex]{graphicx}
\graphicspath{{Figure/pdf/}{Figure/jpeg/}}
\else
\fi
\usepackage[cmex10]{amsmath}
\usepackage{dsfont}
\usepackage{hyperref}
\usepackage{subfigure}
\usepackage{url}

\usepackage{enumitem}

\begin{document}
%

\title{Panoptic Studio: A Massively Multiview System for Social Interaction Capture}
%
%
%

\author{Hanbyul Joo,
        Tomas Simon,
        Xulong Li, Hao Liu, Lei Tan, Lin Gui, Sean Banerjee, Timothy Godisart,\\
        Bart Nabbe, Iain Matthews, Takeo Kanade, Shohei Nobuhara, and Yaser Sheikh
        	\IEEEcompsocitemizethanks{
        		\IEEEcompsocthanksitem H. Joo, T. Simon, B. Nabbe, I. Matthews, T. Kanade, and Y. Sheikh are with Robotics Institute, Carnegie Mellon University, USA. \protect\\
        		Email: \{hanbyulj, tsimon, bana, iainm, tk, yaser\}@cs.cmu.edu. 
        		         		\IEEEcompsocthanksitem X. Li is with Beijing University of Posts and Telecommunications, China. \protect\\
        		         		Email: lixulong@bupt.edu.cn. 
        		         		\IEEEcompsocthanksitem H. Liu and L. Gui are with Ocean University of China, China. \protect\\
        		         		Email: liu.hao@ouc.edu.cn and lgui@qnlm.ac
        		         		\IEEEcompsocthanksitem L. Tan is with Hunan University, China. \protect\\
        		         		Email: leit@hnu.edu.cn. 
								\IEEEcompsocthanksitem S. Banerjee is with Clarkson University, USA. \protect\\
     		         		    Email: sbanerje@clarkson.edu.
        		         		\IEEEcompsocthanksitem T. Godisart is with Oculus Research Pittsburgh, USA. \protect\\
        		         		Email: Timothy.godisart@oculus.com.
        		         		\IEEEcompsocthanksitem S. Nobuhara is with Kyoto University, Japan. \protect\\
        		         		Email: nob@i.kyoto-u.ac.jp. 
        		}
        	}

\IEEEtitleabstractindextext{%
	\begin{abstract}
		We present an approach to capture the 3D  motion of a group of people engaged in a social interaction. The core challenges in capturing social interactions are: (1) occlusion is functional and frequent; (2) subtle motion needs to be measured over a space large enough to host a social group; (3) human appearance and configuration variation is immense; and (4) attaching markers to the body may prime the nature of interactions. The Panoptic Studio is a system organized around the thesis that social interactions should be measured through the integration of perceptual analyses over a large variety of view points. We present a modularized system designed around this principle, consisting of integrated structural, hardware, and software innovations. The system takes, as input, 480 synchronized video streams of multiple people engaged in social activities, and produces, as output, the labeled time-varying 3D structure of anatomical landmarks on individuals in the space. Our algorithm is designed to fuse the ``weak'' perceptual processes in the large number of views by progressively generating skeletal proposals from low-level appearance cues, and a framework for temporal refinement is also presented by associating body parts to reconstructed dense 3D trajectory stream. Our system and method are the first in reconstructing full body motion of more than five people engaged in social interactions without using markers. We also empirically demonstrate the impact of the number of views in achieving this goal. 

	\end{abstract}
	
	\begin{IEEEkeywords}
			\url{https://domedb.perception.cs.cmu.edu}
	\end{IEEEkeywords}}
		

\maketitle


\IEEEdisplaynontitleabstractindextext

%
\IEEEpeerreviewmaketitle


\begin{figure*}[t]
	
	\includegraphics[width=\linewidth]{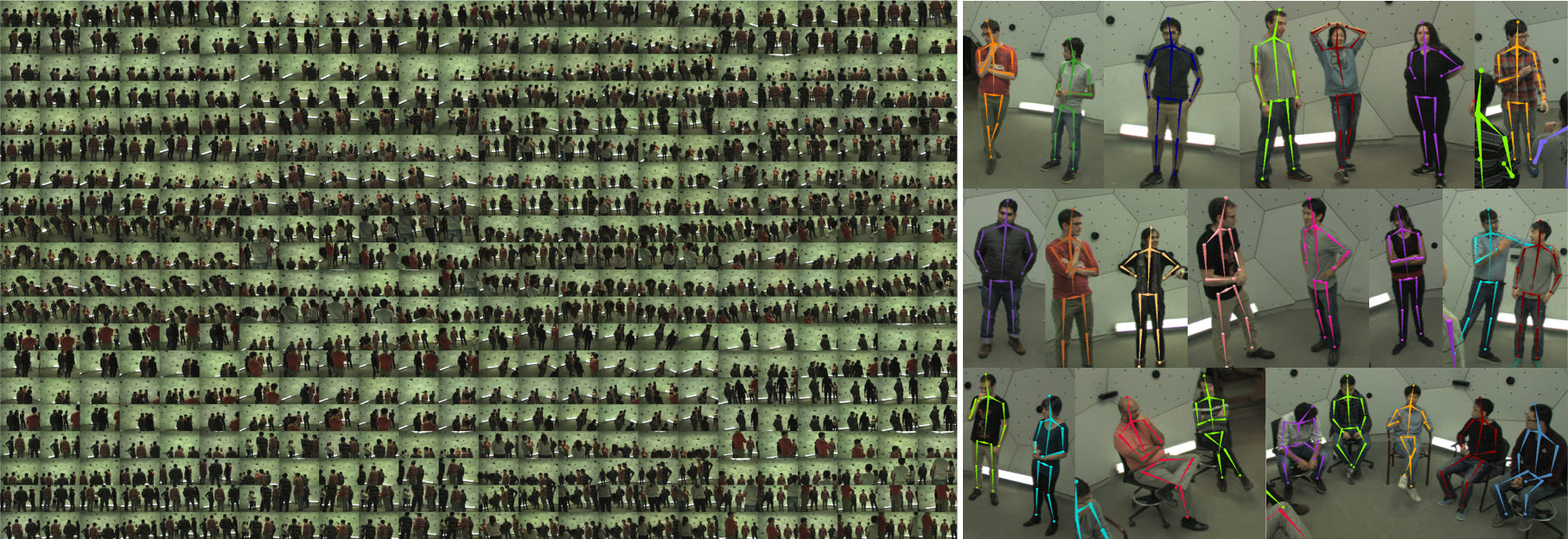}
	\caption{(Left) 480 unique VGA views capturing a social interaction within the Panoptic Studio. (Right) HD example views showing frequently occurring postures that carry rich social signals, with 3D body pose automatically annotated by our method.} 
	\label{fig:iconicPoses}
\end{figure*}

\IEEEraisesectionheading{\section{Introduction}\label{sec:introduction}}

\IEEEPARstart{D}{espite} the fundamental role nonverbal cues play in enabling social function~\cite{Birdwhistell-1970,Philpott-1983}, the protocol underlying this communication is poorly understood---Sapir~\cite{Sapir-1949} called it ``an elaborate code that is written nowhere, known to no one, and understood by all". Some structures of this code have been identified through observational study, such as reciprocity~\cite{Brazelton-1974} or synchrony~\cite{Condon-1974}. However, systematic studies of such phenomena have remained almost entirely focused on the analysis of facial expressions, despite emerging evidence~\cite{Meeren-2005,Aviezer-2012} that facial expressions provide a fundamentally \emph{incomplete} characterization of nonverbal communication. One proximal cause for this singular focus on the face is that capturing natural social interaction presents challenges that current state-of-the-art motion capture systems simply cannot address. This paper describes an approach to capture social signals in natural human interactions, presenting fundamental innovations that span capture design architecture, motion reconstruction algorithms, and a large scale dataset capturing more than 3 hours of group interaction scenes using 521 heterogeneous sensors.

There are four principal challenges in capturing social signaling between individuals in a group: (1) social interactions have to be measured over a volume sufficient to house a dynamic social group, yet subtle details of the motion where important social signals are embedded must be captured; (2) strong occlusions emerge functionally in natural social interactions (e.g., people systematically face each other while interacting, bodies are occluded by gesticulating limbs); (3) human appearance and configuration variation is immense; and (4) social signaling is sensitive to interference---for instance, attaching markers to the face or body, a pre-capture model building stage, or even instructing each individual to assume a canonical body pose during an interaction, primes the nature of subsequent interactions. 

In this paper, we present a system designed to address these issues, with integrated innovations in hardware design, motion representation, and motion reconstruction. The organizing principle is that social motion capture should be performed by the consolidation of a large number of ``weak" perceptual processes rather than the analysis of a few sophisticated sensors. The large number of views provide robustness to occlusions, provide precision over the capture space, and facilitate the boosting of weak 2D human pose detectors into a strong 3D skeletal tracker without any prior about the scenes and subjects. In particular, our contributions include: 

1) \textbf{Modularized Hardware}: We present the modular design of a massively multiview capture consisting of 480 VGA cameras, 31 HD Cameras, and 10 Kinect v2 RGB+D sensors, distributed over the surface of geodesic sphere with a 5.49m diameter (sufficient to house social groups).  

2) \textbf{3D Motion Reconstruction Algorithm for Interaction Capture}:  We present a method to automatically reconstruct full body motion of interacting multiple people. Our method does not rely on a 3D template model or any subject-specific assumption such as body shape, color, height, and body topology. Yet, our method works robustly in various challenging social interaction scenes of arbitrary number of people, producing temporally coherent time-varying body structures. Furthermore, our method is free from error accumulation and, thus, enables capture of long term group interactions (e.g., more than 10 minutes). 


3) \textbf{Social Interaction Dataset}: We publicly share a novel dataset which is the largest in terms of the number of views (521 views), duration (3+ hours in total), and the number of subjects in the scenes (up to 8 subjects) for full body motion capture. Our dataset is distinctive from the previously presented datasets in that ours captures natural interactions of groups without controlling their behavior and appearance, and contains motions with rich social signals as shown in Figure~\ref{fig:iconicPoses} (right). The system described in this paper provides empirical data of unprecedented resolution with the promise of facilitating data-driven exploration of scientific conjectures about the communication code of social behavior. All the data and output are publicly shared on our website\footnote{ \url{https://domedb.perception.cs.cmu.edu}}.

\section{Related Work}

\subsection{Automated Group Behavior Analysis}
Over the last decade, there has been increasing interest in automatically analyzing multiple people's social interaction using multiple camera sensors. Several datasets recording unstructured social scenes are presented, where multiple people (from 5 to 14 subjects) naturally communicate without restriction in their behavior~\cite{Zen-10,Cristani-11,SALSA-15}. In contrast to the scenes captured in structured environment such as round-table meetings~\cite{Lepri-12}, the subjects in such unstructured environments show richer social signals in their body motion, locations, and orientations. However, due to the unconstrained nature, it is challenging to measure their body motion because of severe occlusions among people. Thus, the previous work in this area usually aims to get coarse measurements (e.g., quantized body/head orientation), and they rather focus on higher level social understanding from the coarse measurements, such as F-formation detection~\cite{Lepri-12} and personality predictions~\cite{SALSA-15,Zen-10}. None of the previous work in this domain addresses reconstructing full body skeletal motion of individuals in such challenging scenarios, although rich social signals are embedded in those subtle motions.

\subsection{Markerless Motion Capturing Using Multiple View Systems}


In computer vision, there has been a large number of approaches to measure 3D structure and motion of dynamically moving people using multiple camera sensors. Kanade et al.~\cite{Kanade-1997} pioneered the use of multi-view sensing systems to ``virtualize" reality, using 51 cameras mounted on a geodesic dome of 5 meters in diameter. A number of systems were subsequently proposed to produce realtime virtualizations~\cite{Matusik-2000,Matsuyama-2002,Gross-2003,Petit-2009}. Vlasic et al.~\cite{Vlasic-2009} recovered detail by applying multi-view photometric stereo constraints using a system with 1200 lights on a dome and eight cameras. More recently, a multimodal multi-view stereo system fusing 53 RGB cameras and 53 infrared cameras has been proposed to reconstruct high quality 3D virtual characters~\cite{Collet-15}. 

Other methods explicitly tackle the markerless motion capture by producing 3D skeletal structures over time similar to the marker-based counterparts~\cite{Gall-09, Gavrila-96, Cheung-05,Plankers-03, Bregler-04, Kehl-06, Corazza-10, Vlasic-08, Brox-10, Stoll-11, deAguiar-2008, Vlasic-2008, Furukawa-2008}. The methods deform pre-defined articulated templates of fixed topology to recover the details that were subsampled or occluded in the set of views at a time instant. These methods require an offline method to generate a rigged 3D model for each individual, and the quality of the template is important to achieve high accuracy. The template models need to be aligned at the initial frame to be tracked, and usually a predefined pose (such as a T-pose) are assumed and performed by all individuals. The methods in this area fundamentally suffer from topological changes restricted by the template model, and, similar to other tracking methods, error accumulation is a critical issue in tracking for long durations. Although the 3D template-based method shows good performance---and has become a standard in markerless motion capture approaches---the requirement of a high quality 3D template for each individual limits the practicality of the method, especially in our scenario where dozens of individuals are involved, as the method does not scale well to multiple people. Previous work is demonstrated on a single actor with few exceptions \cite{Ye-2012, Liu-2013}. For example, it is required to segment image cues per subject to track them independently as in~\cite{Liu-2013}, which becomes more complicated if a large number of people are involved, as in our scenes. It should be noted that none of the previous markerless motion capture approaches focus on capturing non-verbal social behaviors of naturally interacting multiple people.

\subsection{Pose Detection Based Approach}


Over the last few years, single view 2D pose estimation method shows great advances based on Convolutional Neural Network framework with large scale human pose datasets~\cite{Tompson-14,Wei-2016}. The state-of-the-art method~\cite{Wei-2016} shows an excellent performance in various environments with varying subject's shape, appearance, and scales. Recently, a few methods facilitate body pose detectors in multiple views to reconstruct 3D body poses~\cite{Burenius2013, Amin-13, Belagiannis2014, Elhayek-15, Elhayek-16}. To infer 3D skeletal parameters from 2D pose detection cues, unary and pairwise terms are defined based on the pre-training data of joint length, relative joint angles, and body colors. The methods are performed at each time independently in fewer camera settings, and thus they typically suffer from motion jitter. Although the results show potential in general environment settings (e.g., outdoors), the methods in this category do not yet reach similar quality compared to the 3D template-based approaches. 





\section{Modularized Hardware Design}
For social motion capture, we design a massively multiview system with heterogeneous sensors including 480 VGA cameras, 31 HD cameras, 10 Kinects. 
The large number of cameras at unique viewpoints provide a large volume with robustness against occlusions, and allow no restriction for view direction of the subjects. The HD views provide details for the scene. Multiple Kinects provide initial point clouds to a generate dense trajectory stream. 



\subsection{Structural Design}
\begin{figure}
	\centering       
	\includegraphics[trim=0 0 0 0,clip,width=0.95\linewidth]{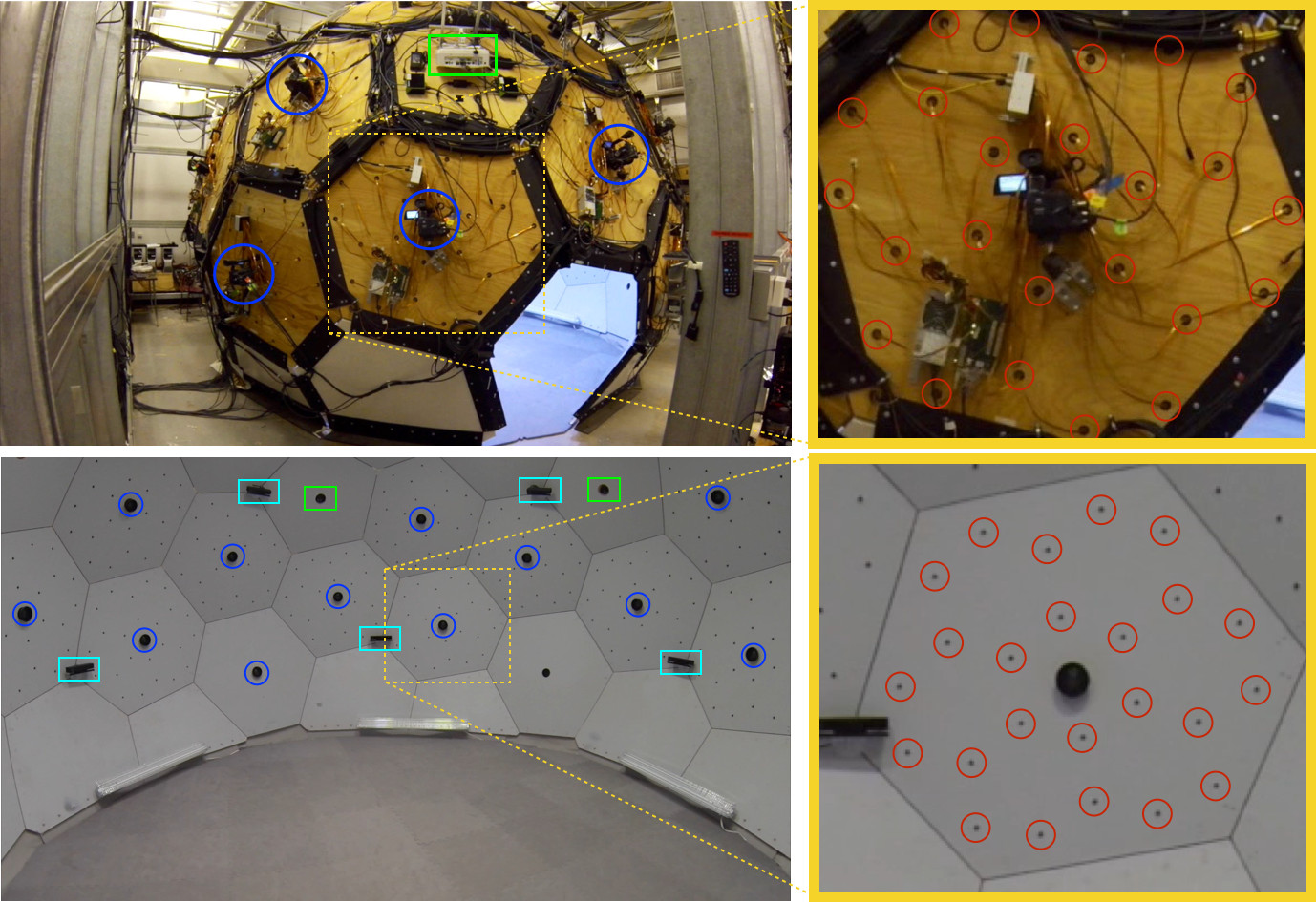}
	\includegraphics[trim=0 0 100 0,clip,width=0.45\linewidth]{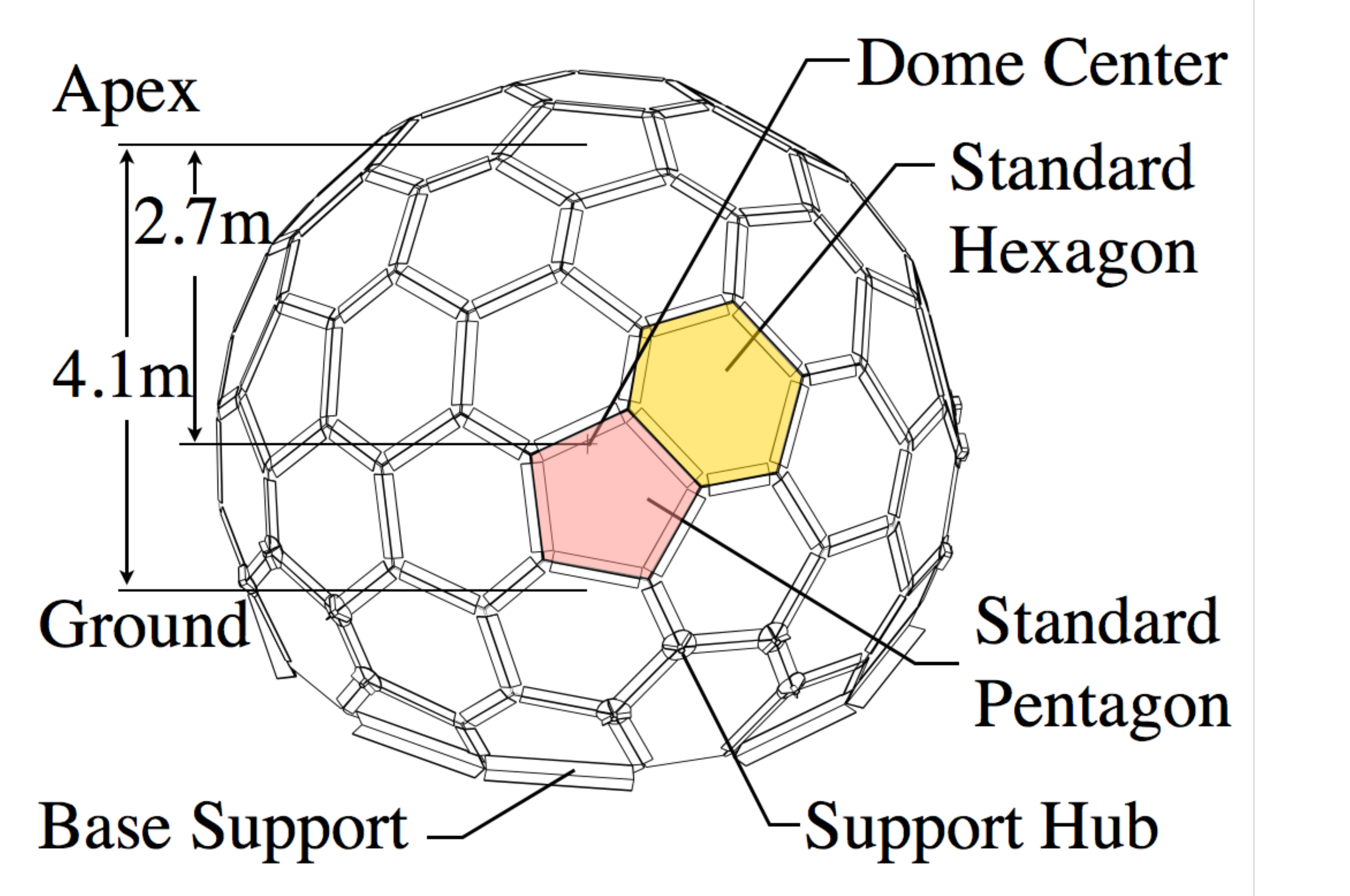} 
	\includegraphics[trim=230 50 400 50,clip,width=0.45\linewidth]{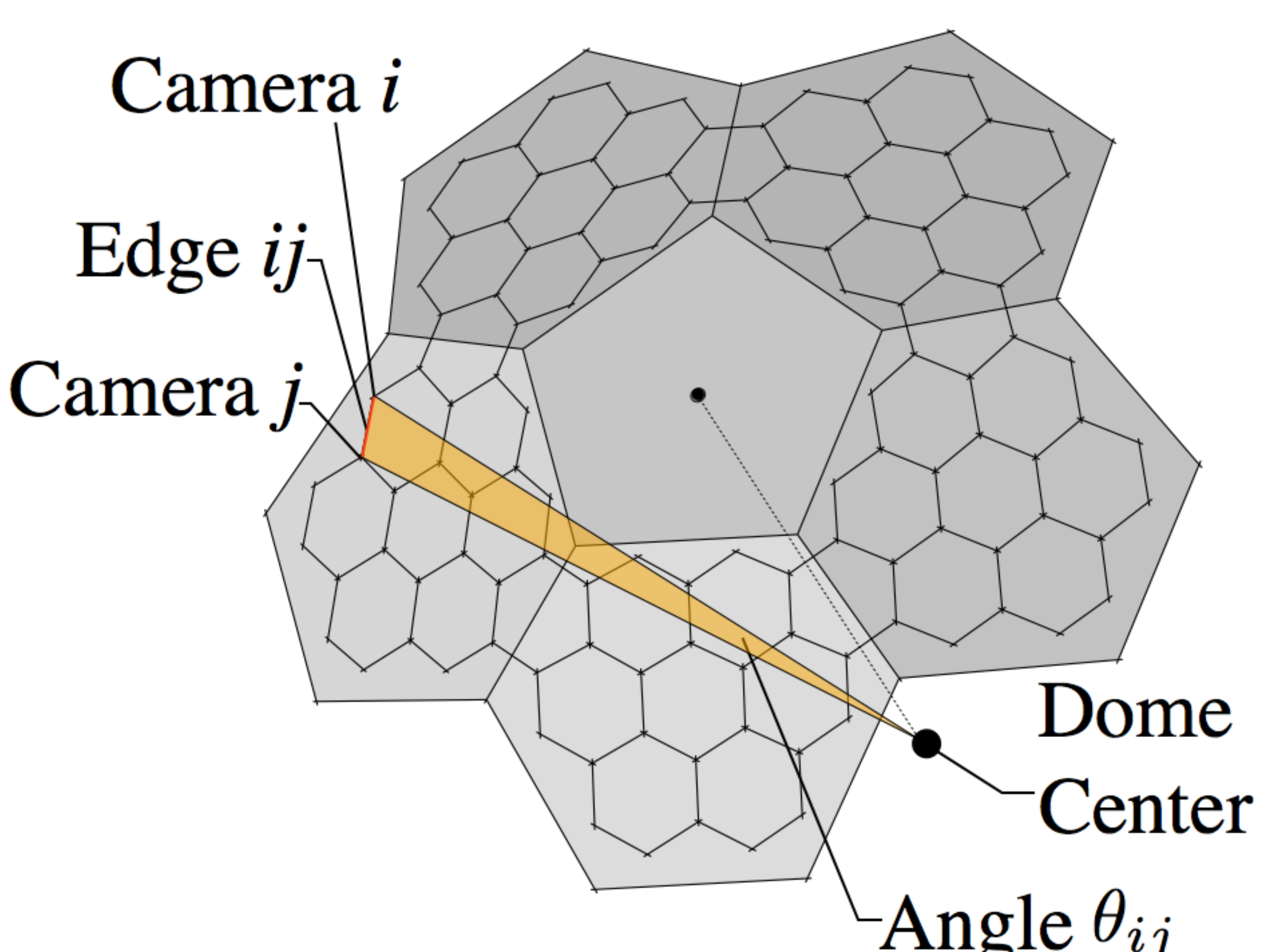}
	\caption{The studio structure. (Top) The exterior of the dome with the equipment mounted on the surface. (Middle) The interior of the dome. VGA cameras are shown as red circles, HD cameras as blue circles, Kinects as cyan rectangles, and projectors as green rectangles. (Bottom left) The panels are designed to ensure interchangeability. (Bottom right) Optimized camera positions to ensure uniform angles with respect to the dome center between each camera and all its neighbors (e.g., Camera $i$ is a neighbor of Camera $j$).} 
	\label{fig:domeFigure}
\end{figure}

The physical frame of the studio is a variant of a face-transitive solid called a truncated pentagonal hexecontahedron. This particular structure was selected because it has among the largest number of transitive faces of any geodesic dome~\cite{Williams1979}. The transitivity of the faces enables the modular architecture, and ensures that the structure remains easy to upgrade and customize with different panels of the same configuration. The structure has a diameter of 5.49m and a total height of 4.15m. The floor of the dome is 1.40m below the center to increase access to the edges, as shown in Figure~\ref{fig:domeFigure}. In all, the structure consists of 6 pentagonal panels, 40 hexagonal panels, and 10 trimmed base panels. 

Our design was modularized so that each hexagonal panel houses a set of 24 VGA cameras. To determine the placement of the VGA cameras, we initialized their positions by tessellating the hexagon face into 24 triangles and using this initialization to define a 3-neighborhood structure shown in the bottom right illustration of Figure~\ref{fig:domeFigure}. Using this neighborhood structure and the initialization we determine the placement of the cameras over the geodesic dome by minimizing the difference in angles between all neighbors of every camera,
{\small
	\begin{equation}\nonumber
	\{\theta_{ij}\}^* = \arg \min_{\{\theta_{ij}\}} \sum_{p=1}^P \sum_{i=1}^{N} \sum_{j \in \mathcal{N}(i)}  \sum_{k \in \mathcal{N}(i) \neq j}  (r(\theta_{ij}|p)-r(\theta_{ik}|p))^2 ,
	\end{equation}
}where $P=20$ is the number of panels, $N=24$ is the number of cameras in each panel, $\mathcal{N}(\cdot)$ is the neighborhood of a camera, $r(\cdot|p)$ is a function transforming the angle on a reference panel to the $p$-th panel. The cameras sample the span of the vertical axis of the space and sample $48.71^\circ$ of the horizontal axis. With this distribution, the minimum baseline between any VGA camera and its nearest three neighbors is 21.05cm.

The 31 HD cameras are installed at the center of each hexagonal panel, and 5 projectors are installed at the center of each pentagonal panel\footnote{Note that no sensors are installed on some panels (e.g., ceiling panels occluded by lights).}. Additionally, a total of 10 Kinect v2 RGB+D sensors are mounted at heights of 1 and 2.6 meters, forming two rings with 5 evenly spaced sensors each. The interior and exterior of our system are shown in Figure~\ref{fig:domeFigure}. 

\subsection{System Architecture}
\begin{figure}[t]
	\includegraphics[width=\linewidth]{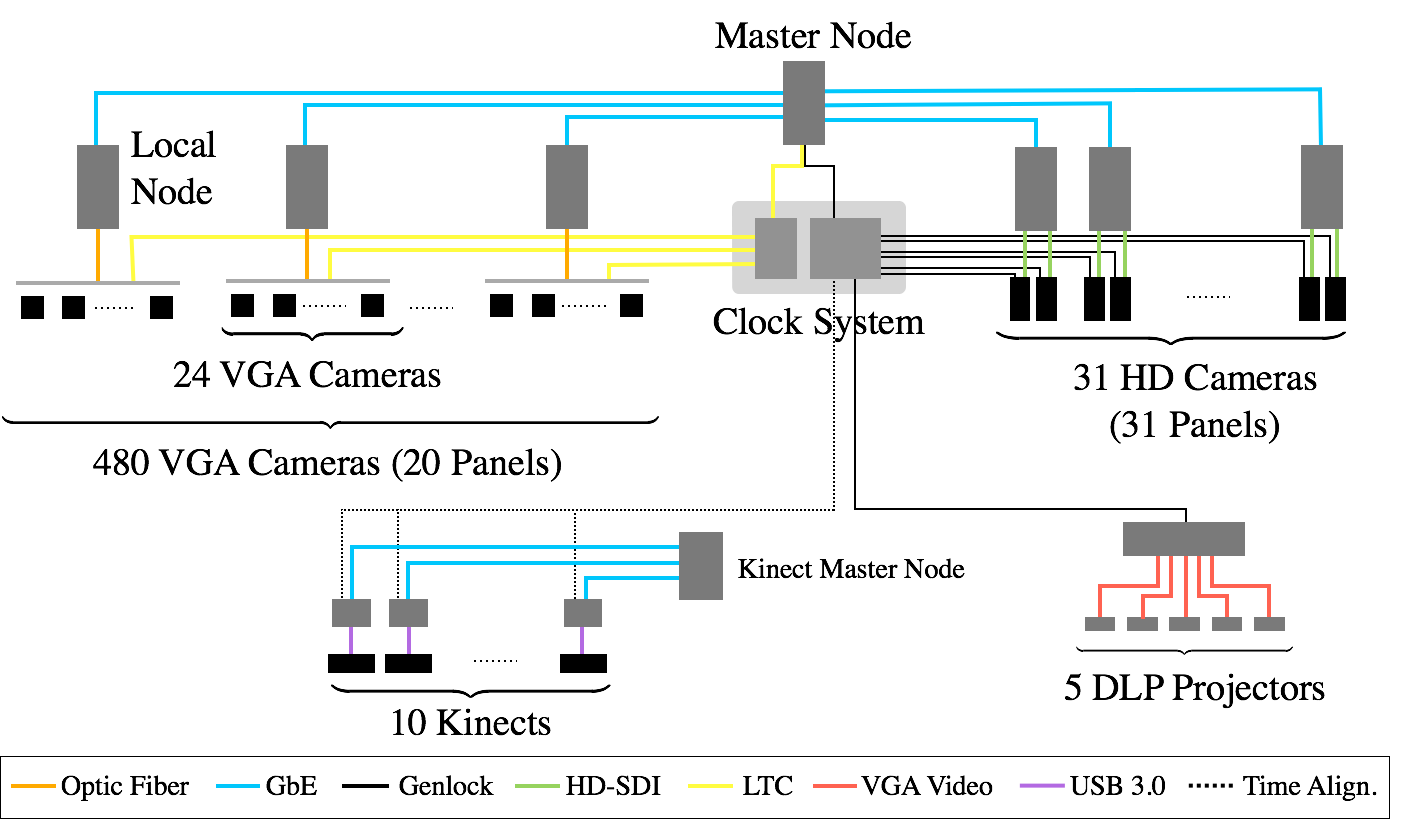}
	\caption{Modularized system architecture. The studio houses 480 VGA cameras synchronized to a central clock system and controlled by a master node. 31 synchronized HD cameras are also installed with another clock system. The VGA clock and HD clock are temporally aligned by recording them as a stereo signal. 10 RGB-D sensors are also located in the studio. All the sensors are calibrated to the same coordinate system.}
	\label{fig:architecture}
\end{figure}

Figure \ref{fig:architecture} shows the architecture of our system. The 480 cameras are arranged modularly with 24 cameras in each of 20 standard hexagonal panels on the dome. Each module in each panel is managed by a Distributed Module Controller (DMC) that triggers all cameras in the module, receives data from them, and consolidates the video for transmission to the local machine. Each individual camera is a global shutter CMOS sensor, with a fixed focal length of 4.5mm, that captures VGA ($640\times480$) resolution images at 25Hz. 

Each panel produces an uncompressed video stream at 1.47 Gbps, and thus, for the entire set of 480 cameras the data-rate is approximately 29.4 Gbps. To handle this stream, the system pipeline has been designed with a modularized communication and control structure. For each subsystem, the clock generator sends a frame counter, trigger signal, and the pixel clock signal to each DMC associated with a panel. The DMC uses this timing information to initiate and synchronize capture of all cameras within the module. Upon trigger and exposure, each of the 24 camera heads transfers back image data via the camera interconnect to the DMC, which consolidates the image data and timing from all cameras. This composite data is then transferred via optical interconnect to the module node, where it is stored locally. Each module node has dual purpose: it serves as a distributed RAID storage unit\footnote{Each module has 3 HDDs integrated as RAID-0 to have sufficient write speed without data loss, totaling 60 HDDs for 20 modules.} and participates as a multi-core computational node in a cluster. All the local nodes of our system are on a local network on a gigabit switch. The acquisition is controlled via a master node that the system operator can use to control all functions of the studio.

Similar to the VGA cameras, HD cameras are modularized and each pair of cameras are connected to a local node machine via SDI cables. Each local node saves the data from two cameras to two RAID storage units respectively.


Each RGB+D sensor is connected to a dedicated capture node that is mounted on the dome exterior. To capture at rates of approximately 30 Hz, the nodes are equipped with two SSD drives each and store color, depth, and infrared frames as well as body and face detections from the Kinect SDK. A separate master node controls and coordinates the 10 capture nodes via the local network.

\subsection{Temporal Calibration for Heterogeneous Sensors}
Synchronizing the cameras is necessary to use geometric constraints (such as triangulation) across multiple views. In our system, we use hardware clocks to trigger cameras at the same time. Because the frame rates of the VGA and HD cameras are different (25 fps and 29.97 fps respectively) we use two separate hardware clocks to achieve shutter-level synchronization among all VGA cameras, and independently among all HD cameras. To precisely align the two time references, we record the timecode signals generated from the two clocks as a single stereo audio signal, which we then decode to obtain a precise alignment at sub-millisecond accuracy. 

Time alignment with the Kinect v2 streams (RGB and depth) is achieved with a small hardware modification: each Kinect's microphone array is rewired to instead record an LTC timecode signal\footnote{As a result of this modification, microphone output on the Kinects is therefore discarded. More details about this hardware modification are available upon request.}.  This timecode signal is the same that is produced by the genlock and timecode generator used to synchronize the HD cameras, and is distributed to each Kinect via a distribution amplifier. We process the Kinect audio to decode the LTC timecode, yielding temporal alignment between the recorded Kinect data---which is timestamped by the capture API for accurate relative timing between color, depth, and audio frames---and the HD video frames. Empirically, we have confirmed the temporal alignment obtained by this method to be of at least millisecond accuracy.

\subsection{Spatial Calibration}
We use Structure from Motion (SfM) to calibrate all of the 521 cameras. To easily generate feature points for SfM, five projectors are also installed on the geodesic dome. For calibration, they project a random pattern on a white structure (we use a portable white tent), and multiple scenes (typically three) are captured by moving the structure within the dome. We perform SfM for each scene separately and perform a bundle adjustment by merging all the matches from each scene. We use the VisualSfM software~\cite{vsfm} with 1 distortion parameter to produce an initial estimate and a set of candidate correspondences, and subsequently run our own bundle adjustment implementation with 5 distortion parameters for the final refinement. The computation time is about 12 hours with 6 scenes (521 images for each) using a 6 core machine. 
In this calibration process, we only use the color cameras of Kinects. We additionally calibrate the transformation between the color and depth sensor for each Kinect with a standard checkerboard pattern, placing all cameras in alignment within a global coordinate frame.

%
%

\begin{figure*}[t!]
	\includegraphics[width=\linewidth]{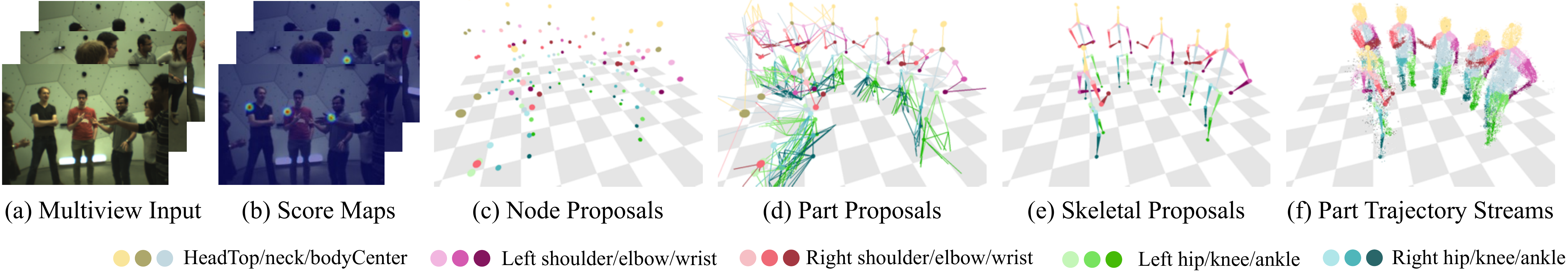}
	\caption{Several levels of proposals generated by our method. (a) Images from upto 480 views. (b)~Per-joint detection score maps. (c)~Node proposals generated after non-maxima suppression. (d)~Part proposals by connecting a pair of node proposals. (e)~Skeletal  proposals generated by piecing together part proposals. (f)~Labeled 3D patch trajectory stream showing associations with each part trajectory. In~(c-f), color means joint or part labels shown below the figure.} 
	\label{fig:overview}
\end{figure*}

\begin{table}[t]	
	\renewcommand{\arraystretch}{1.3}
	\caption{Summary of Notation.}
	\label{Table:notations}
	\centering
	\begin{tabular}{l|l}
		\hline 
		Notation & Descriptions \tabularnewline
		\hline 	
		$s_{i}^c $ & $i$-th 2D skeleton detection in a camera view $c$ \tabularnewline		
		\hline 
		$s_{ij}^c $ & $j$-th joint of $i$-th 2D skeleton in a camera view $c$  \tabularnewline		
		\hline 
		$h_{ij}^{c}(\mathbf{z})$ & 2D score map of $j$-th joint of $i$th skeleton in view $c$  \tabularnewline		
		\hline 
		$h_{j}^{c}(\mathbf{z}) $ & Merged score map of $j$-th joint of all skeletons in view $c$  \tabularnewline		
		\hline 
		${H}_{j}(\mathbf{Z})$ & 3D score map for the $j$-th joint \tabularnewline		
		\hline 
		$\mathbf{N}_j$ & Node proposals for the $j$-th joint\tabularnewline		
		\hline 
		$\mathbf{P}_{uv}$ & Part proposals for the part connecting nodes $(u,v)$ \tabularnewline		
		\hline 
		$\mathbf{S}$ & Skeletal proposals connecting multiple part proposals \tabularnewline		
		\hline 
		$\mathbf{\tilde{S}}(t)$ & Skeletal trajectory proposals, associated through time  \tabularnewline		
		\hline   
		$\mathbf{\tilde{P}}_{uv}(t)$ & Part trajectory proposals for the connecting nodes $(u,v)$ \tabularnewline		
		\hline 
		$\mathbf{F}$	& 3D Patch Trajectory Stream, $\{\mathbf{f}_i\}_{i=1}^{N_F}$  \tabularnewline		
		\hline   
		$\mathbf{F}_{uv}$	& A subset of  $\mathbf{F}$ associated to $\mathbf{\tilde{P}}_{uv}$ \tabularnewline		
		\hline   
	\end{tabular} 
\end{table}

\section{Method Overview and Notation}
Our algorithm is composed of two major stages. The first stage takes, as input, images from multiple views at a time instance (calibrated and synchronized), and produces 3D body skeletal proposals for multiple human subjects. The second stage further refines the output of the first stage by using a dense 3D patch trajectory stream~\cite{Joo2014}, and produces temporally stable 3D skeletons and an associated set of labeled 3D patch trajectories for each body part, describing subtle surface motions.

In the first stage, a 2D pose detector~\cite{Wei-2016} is computed independently on all 480 VGA views at each time instant $t$, generating detection score maps for each body joint (see Fig.~\ref{fig:overview}b). The 2D score maps for each body joint~$j\in\{1,\cdots,J\}$ are combined into a 3D score map $H_j(\mathbf{Z})$ by projecting a grid of voxels $\mathbf{Z}\in \mathds{R}^3$ onto the 2D score maps and computing an average 3D score at each voxel (subsection~\ref{subsection:nodeProposals}).

Our approach then generates several levels of proposals, as shown in Figure~\ref{fig:overview}. A set of \textit{node proposals} $\mathbf{N}_j$ for each joint $j$ is generated by non-maxima suppression of the 3D score~map~$H_j(\mathbf{Z})$, where the $k$-th node proposal~$\mathbf{N}_j^k \in \mathds{R}^{3}$ is a putative 3D position of that anatomical landmark. Similarly, the set of \textit{part proposals} is denoted by $\mathbf{P}_{uv}$, where $u$ and $v$ are joints and $(u,v) \in \mathbf{B}$ is the set of body parts or {\em bones} composing a skeleton hierarchy.
The $k$-th part proposal, $\mathbf{P}_{uv}^k= (\mathbf{N}_u^{k_u}, \mathbf{N}_v^{k_v}) \in \mathds{R}^6$, is a putative body part connecting two node proposals, $\mathbf{N}_u^{k_u}$ and $\mathbf{N}_v^{k_v}$, where the index $k$ enumerates all possible combinations of $k_u$ and $k_v$. As the output of the first stage, our algorithm produces \textit{skeletal proposals}; we refer to the $k$-th proposal as $\mathbf{S}^k = \{\mathbf{P}_{uv}^{k}\}_{uv \in \mathbf{B}}$. A skeletal proposal is generated by finding an optimal combination of part proposals using a dynamic programming method under the score function defined in subsection~\ref{subsection:dynamicProgamming}. Here, we abuse the notation to have $\mathbf{P}_{uv}^{k}$ refer to the optimally assigned part $u,v$ of skeleton $k$ (the superscript $k$ is understood to be the optimal mapping, from context). After reconstructing skeletal proposals at each time $t$ independently, we associate skeletons from the same identities across time and generate \textit{skeletal trajectory proposals} $\mathbf{\tilde{S}}^k(t) = \{\mathbf{\tilde{P}}^k_{uv}(t)\}_{uv \in \mathbf{B}}$, where $\mathbf{\tilde{P}}^k_{uv}(t)$ is a \textit{part trajectory proposal}, a moving part across time, with $k$ similarly overloaded to denote the optimal associations determined in each frame $t$.


In the second stage, we refine the skeletal trajectory proposals generated in the first stage using dense 3D patch trajectories~\cite{Joo2014}. To produce evidence of the motion of different anatomical landmarks, we compute a set of dense 3D trajectories $\mathbf{F}=\{\mathbf{f}_i\}_{i=1}^{N_F}$, which we refer to as a \textit{3D patch trajectory stream}, by tracking each 3D patch independently. Each patch trajectory $f_i$ is initiated at an arbitrary time (every 20th frame in our results), and tracked for an arbitrary duration (30 frames backward-forward in our results) using the method of Joo et al. \cite{Joo2014}. Our method associates a part trajectory $\mathbf{\tilde{P}}^k_{uv}$ with a set of patch trajectories $\mathbf{F}_{uv}^k$ out of $\mathbf{F}$, and these trajectories determine rigid transformations, $T(t{+}1\,|\,t) \in SE(3) $, between any time $t$ to $t{+}1$ for this part. These labeled 3D trajectories associated to each part provide surface deformation cues and also play a role in refining the quality by reducing motion jitter, filling missing parts, and detecting erroneous parts.

\begin{figure*}[t]
	\includegraphics[width=\linewidth]{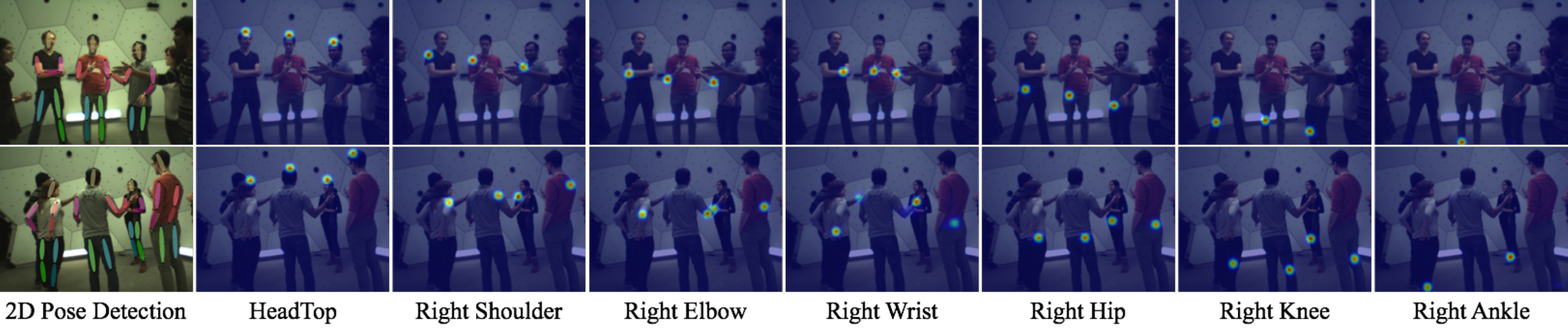}
	\caption{2D pose detections and score maps generated by the method of \cite{Wei-2016}. (Column 1) Example views out of 480 views with proposals by the pose detector (Column 2-7) Heat maps for each node on each view. Note that the body pose detector distinguishes left-right limbs.}\label{fig:poseDetection}
\end{figure*}
\section{The First Stage: Skeletal Proposals Generation}
Our algorithm integrates 2D pose detections across the many views of our massively multiview system, fusing simple 2D cues to estimate 3D skeletal poses at each time instance.
While detections in any single view may be incomplete or inaccurate---typically due to occlusions---we find that aggregating these cues across many views yields very stable results. Our method is simple, yet robust thanks to the large number of views. In contrast, prior marker-less motion capture methods are typically ``model-dependent'', requiring a 3D template model to constrain shape deformations, a motion model to constrain temporal deformations, and a relatively complex energy function minimization that trades off each of these priors~(e.g., \cite{Gall-09,Furukawa-2008, Elhayek-16}). Our method in this stage is essentially based on triangulating detections at a single time instance, and, thus, does not suffer from error accumulation or drift. It does not require a 3D template model, prior assumptions about the subject or the motion, or an initial alignment for tracking. In this section, we describe how the proposals are generated and built up from 2D cues.  


\subsection{3D Node Score Map and Node Proposals}
\label{subsection:nodeProposals}
A single-view 2D pose detector is computed on all VGA views at each time instant, and is used to generate 2D pose detections and per-joint score maps in each image. Because the first stage of our method is performed at each time independently, we will consider a fixed time instant $t$, and drop the time variable for clarity. We use the detector of Wei et al.~\cite{Wei-2016} without additional training. The method of \cite{Wei-2016} requires bounding box proposals for each human body as initialization, thus, we first apply a person detector similar to R-CNN \cite{Girshick-14}, and run the pose detector on the detected person proposals represented as bounding boxes. Each 2D skeleton detection $i$ in a camera view $c$ is denoted by $\mathbf{s}_{i}^c \in \mathds{R}^{2\times 15}$, and is composed of 15 anatomical landmarks or \emph{nodes} (3 for the head/torso and 12 for the limbs), also referred to as joints\footnote{We modify the skeleton hierarchy of \cite{Wei-2016} to have an explicit torso bone, by taking the center of the two hip nodes as a body center node.}. The position of the $j$-th node of the $i$-th person detection is denoted by $\mathbf{s}_{ij}^c \in \mathds{R}^{2}$. The method of \cite{Wei-2016} also provides a score map representing the per-pixel detection confidence for each node $s_{ij}^c$, which we denote as $h_{ij}^{c}(\mathbf{z})\in[0,1]$, where $\mathbf{z}\in \mathds{R}^2$ indexes 2D image space. We also compute a merged score map by taking the maximum across all person detections at each pixel, $h_{j}^{c}(\mathbf{z})=\max_i h_{ij}^{c}(\mathbf{z})$. Merged score maps of example views are shown in Figure~\ref{fig:poseDetection}.

To combine 2D node score maps from multiple views into 3D, we generate a 3D score map for each node using a spatial voting method. We first index the 3D working space into a voxel grid (4cm in our implementation), and compute the \textit{node-likelihood} score of each voxel by projecting the center of the voxel to all views and taking the average of the 2D scores at the projected locations. The 3D score map $H_j(\mathbf{Z})$ for a node $j$ at the 3D position $\mathbf{Z}\in\mathds{R}^3$ is defined as 
\begin{equation}
{H}_j(\mathbf{Z}) = \frac{1}{ | V(\mathbf{Z}) | } \sum_{c \in V(\mathbf{Z})} {h^c_j \left(\mathcal{P}_c({\mathbf{Z}}) \right) }  ,
\end{equation}
where $\mathcal{P}_c(\cdot)\in\mathds{R}^2$ denotes projection into camera~$c$, $V(\mathbf{Z})$ is the set of cameras where the 3D location $\mathbf{Z}$ is visible, and $| V(\mathbf{Z}) |$ is the cardinality of the set. Note that the 3D score map for each node is computed separately, producing fifteen 3D score maps at each time instant. 

From the 3D score map for each node at each time instance, we perform Non-Maxima Suppression (NMS), and keep all the candidates above a fixed threshold $\tau$ (we use $\tau{=}0.05$). The results are shown in the Figure~\ref{fig:overview}c, and the same results color-coded by the node scores are shown in Figure~\ref{fig:proposalScores}. Each node proposal, denoted as $\mathbf{N}_j^k$ for the $k$-th proposal for node $j$, is a putative candidate for the $j$-th anatomical landmark of a participant.


\subsection{Part Proposals}
\label{subsection:partProposals}

\begin{figure}
	\includegraphics[width=\linewidth]{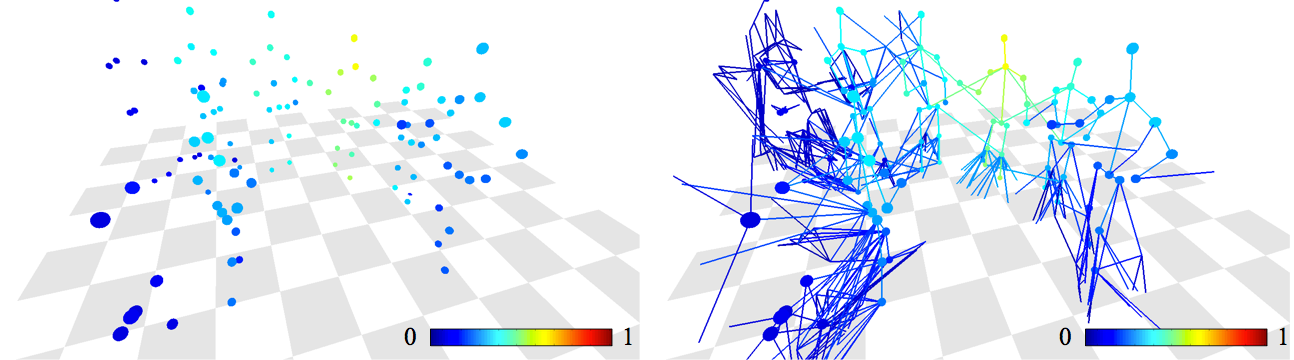}
	\caption{Computed scores for node proposals and part proposals. The color encodes scores.} 
	\label{fig:proposalScores}
\end{figure}

Given the generated node proposals, we infer part proposals by estimating connectivity between each pair of nodes that make up a possible body part. The 2D detector \cite{Wei-2016} uses appearance information during the inference, and, thus, the result tends to preserve connectivity information (e.g., left knee is connected to the left foot of the same person). Our approach fuses them by voting 2D connectivity into 3D. More specifically, we define a connectivity score between a pair of node proposals by projecting them onto all views, and checking in how many views they are actually connected, i.e., both nodes belong to the same person detection. Formally, the connectivity score of a part $\mathbf{P}_{uv}^k$ between two node proposals $( \mathbf{N}_{u}^{k_u}, \mathbf{N}_{v}^{k_v})$, where $(u,v) \in \mathbf{B}$, is defined as
\begin{gather}
{\Phi}( \mathbf{P}_{uv}^k )=  \frac{1}{ | V( \mathbf{P}_{uv}^k ) | } \sum_{c \in V(\mathbf{P}_{uv}^k)}   \max_{i} \phi_{iuv}^c \left( \mathcal{P}_c(\mathbf{N}^{k_u}_{u}),\mathcal{P}_c(\mathbf{N}^{k_v}_{v}) \right), \nonumber \\
\phi_{iuv}^c(\mathbf{z}_u,\mathbf{z}_v) = w_{iuv}^c(\mathbf{z}_u,\mathbf{z}_v)\delta^c_{iuv} \left(\mathbf{z}_u,\mathbf{z}_v \right) \nonumber
%
%
\end{gather}
where
\begin{align}
w_{iuv}^c(\mathbf{z}_u,\mathbf{z}_v) &= \frac{1}{2}\left(h_{iu}^c(\mathbf{z}_u) + h_{iv}^c(\mathbf{z}_v) \right), \text{and} \nonumber\\
\delta^c_{iuv}( \mathbf{z}_u ,\mathbf{z}_v) &=  \begin{cases}
1 & \mathrm{if}\,\, h_{iu}^c( \mathbf{z}_u ) > \tau ~\mathrm{and}~h_{iv}^c( \mathbf{z}_v ) > \tau\\
0 & {\rm otherwise}.
\end{cases} \nonumber
\end{align}
Here, $\mathcal{P}_c(\mathbf{N}^{k_u}_{u})$ and $\mathcal{P}_c(\mathbf{N}^{k_v}_{v})$ are the projections of the two nodes of $\mathbf{P}_{uv}^k$ in view $c$, and $ V( \mathbf{P}_{uv}^k )$ is the set of cameras where the 3D part is visible. Intuitively, the part score ${\Phi}$ represents the average connectivity score across all views from all potentially corresponding 2D person detections. Because we do not know the correspondence from 3D parts to 2D person detections, we take the maximum score across all possible detections $i$ in each view. Assuming that the projected part corresponds to the $i$-th person detection in camera $c$, the part connectivity score 
$\phi_{iuv}^c$ is defined as the average score of the projected nodes, denoted by $w_{iuv}^c(\mathbf{z}_u,\mathbf{z}_v)$. The delta function $\delta^c_{iuv}$ additionally ensures that $\phi_{iuv}^c$ is nonzero only if both projected node locations have a sufficiently high score for the same detection $i$ (i.e., both nodes are detected as part of a single person). An example of computed part scores is shown in Figure~\ref{fig:proposalScores}.

\subsection{Generating Skeletal Proposals by Dynamic Programming}
\label{subsection:dynamicProgamming}

Our method generates skeleton proposals by piecing together the part proposals. Since each skeleton is a tree structure, this can be computed efficiently using Dynamic Programming (DP)---but only for a single person. Therefore, we use DP to greedily find 3D skeletons $\mathbf{S}^k$ which maximize the sum of part scores,
\begin{gather}
\Theta (\mathbf{S}^k)= \max_{(k_{1},\cdots,k_J)}\sum_{(u,v) \in \mathbf{B}} {\Phi} \left( \mathbf{P}_{uv}^{k} \right). \nonumber
\end{gather}
A skeleton $\mathbf{S}^k$ is given by the mapping $k\mapsto(k_1,\cdots,k_J)$, where the J-tuple $(k_1,\cdots,k_J)$ determines the assignment of node proposals $\mathbf{N}_j^{k_j}$ for each joint $j$ in the body. After picking the highest scoring skeleton $\Theta(\mathbf{S}^k)$, the assigned nodes $(k_1,\cdots,k_J)$ are removed from the pool of available node proposals and we run DP again to find the next highest scoring skeleton, and so on until all possible skeletons are found.

One option here would be to threshold the skeleton scores $\Theta(\mathbf{S}^k)$ at some minimum value to determine valid detections. However, we can do better: each 3D skeleton  should be supported by 2D detections, and each 2D detection can correspond to only a single 3D skeleton. This observation is important because the voting used to generate 3D node proposals assigns equal score to {\em all} voxels along the line of sight of each 2D detection (Sect.~\ref{subsection:nodeProposals}), and, similarly, the $\max$ over detections in the part score $\Phi(\cdot)$ makes  $\Theta(\mathbf{S}^k)$ an overestimate. 

To avoid this form of double counting, our method places each 3D node $\mathbf{N}^{k}_{j}$ in skeleton $\mathbf{S}^k$ in correspondence with the closest 2D joint detection in each view. For each 3D node $\mathbf{N}^{k}_{j}$, we create a set of correspondences $\mathcal{C}_j^k$ with elements $(c,i)$ such that the distance $\| \mathcal{P}_c(\mathbf{N}_j^k) - \mathbf{s}_{ij}^c \|_2$ is the minimum across all detections $i$ in view $c$ and smaller than $\delta{=}10$px. 
Once a 2D correspondence is established, we remove it from the set of available 2D detections, and, as above, this is performed greedily in order of decreasing skeleton score $\Theta(\mathbf{S}^k)$. Skeletons where the head node has fewer than two correspondences are discarded, i.e., if $|\mathcal{C}_{j}^k|<2$ for $j$ the head.

We additionally use the set of correspondences $\mathcal{C}_j^k$ to refine the 3D node locations by minimizing reprojection error. This overcomes the discretization error introduced by the voxel grid resolution. The final 3D node location $\hat{\mathbf{N}}^{k}_{j}$ is then
\begin{gather}
\hat{\mathbf{N}}^k_j = \arg\min_\mathbf{Z} \sum_{(c,i) \in \mathcal{C}_j^k} \left\|\mathcal{P}_c(\mathbf{Z}) - \mathbf{s}_{ij}^c \right\|_2. \nonumber
\end{gather}
The output of the algorithm described in this section is 3D skeletal proposals reconstructed independently at each time instance. After performing this process on all frames, our method associates skeletons from the same identity across time by simply considering spatial distance of the head node. That is, for a $\mathbf{S}_t^{k_1}$ reconstructed at time $t$, we find a corresponding skeleton at $\mathbf{S}_{t+1}^{k_1}$ with the closest head node location from $\mathbf{S}_t^{k_1}$ within a threshold. 
To be somewhat robust to missing skeleton detections, our method associates across a window of time. If there is no corresponding skeleton at time $t+1$, we also consider the next time $t+2$ and find a corresponding skeleton.

This first stage of our method is performed without considering any temporal cues. The advantages of this are that the method can easily handle a varying number of people, there is no need to impose priors on the motion or skeletons, and the bulk of the computation is easily parallelized across frames. In many cases, we find that the results from the first stage are already sufficient for many applications. However, the results exhibit some jitter---especially for complex scenes with limited views per person---and missed or noisy detections do not benefit from evidence found in adjacent frames. We address these issues in the second stage of our method.

\section{The Second Stage: Temporal Refinement and Trajectory Stream Labeling}
The per-frame skeletal proposals from the first stage can be improved by using temporal coherence. We use motion cues from a {\em 3D patch trajectory stream}: dense 3D point tracks computed by the method of Joo et al.~\cite{Joo2014}. We find an association between each part trajectory proposal and a subset of the patches in the trajectory stream, and use it to reduce motion jitter, remove false detections, and fill in missing detections. The resulting labeled patch trajectories also capture rich motion information representing the subtle deformations of the surface for each body part (see Fig.~\ref{fig:overview}f).

%

\subsection{Patch Trajectory Stream Reconstruction}
\label{subsection:patchTrajectoryRecon}
We can only observe surface motions, not the true motion of the underlying skeleton, so it is not immediately apparent how best to enforce temporal consistency in the motion of body parts. Clothing in particular makes the relationship between surface motion and body parts difficult to model. To keep the use of priors and models to a minimum, we therefore choose to measure surface motion independently from the underlying skeletal motion and postpone all decisions about part-to-surface associations.

To represent surface motion, we use the method of \cite{Joo2014} to track a dense 3D patch cloud---a set of points with corresponding surface normal and a small spatial extent, representing the surface locally---and estimate the motion of each of these patches. Instead of generating the initial patches to track using SIFT matching and triangulation (as in \cite{Joo2014}), we use the depth maps from our 10 RGB+D sensors to generate an initial set of 3D patches. For a single frame, a dense 3D point cloud is first generated from the depth maps, and planar local patches centered on each point are initialized. The size of patches is manually determined by considering image resolution and fixed for the entire processes at {6cm$\times$6cm}. To find the normal of each patch, we apply Singular Value Decomposition (SVD) to the coordinates of points within a neighborhood (determined by Euclidean distance from the center point with the patch size as a threshold), and the least principal axis is selected as the normal direction. The sign of the normal is disambiguated by considering camera visibility. 

The remainder of the algorithm (3D patch tracking) is as described in~\cite{Joo2014}. As a brief overview, a patch is represented by a triplet points (the center point, and two orthogonal points on the patch plane), and it is tracked by projecting the triplet points on all views where the target patch is visible. Optical flow tracking is performed in 2D on each point, and the tracked 2D flows are triangulated into 3D. The core idea to fully leverage a large number of views is to reason about the time-varying camera visibility of each patch. The visibility is optimally estimated in a MAP framework that combines photometric consistency, motion consistency, and visibility priors, see \cite{Joo2014} for more details. For our results, we initialize a 3D patch cloud every 20th frame, and track them backward and forward for 30 frames in each direction. As output, we obtain a dense 3D patch trajectory stream, $\mathbf{F}=\{\mathbf{f}_i\}_{i=1}^{N_F}$, where each $\mathbf{f}_i(t)\in\mathds{R}^3$ is the time-varying position of a tracked patch.

\subsection{Associating Part Trajectory Proposals and Trajectory Stream}
\label{subsection:trajectoryAssociation}
Part trajectory proposals $\mathbf{\tilde{P}}_{uv}$ represent the moving body parts of a single person, and are given by the optimal assignment used to generate skeletal trajectory proposals. These part trajectories lack temporal coherence because they are reconstructed independently in each frame. However, the trajectory streams provide evidence of the motion of each limb, and can be used to refine the motion of each body part. We therefore find an association between each part trajectory proposal and a subset of patch trajectories. This can be seen as a semantic labeling of the patch trajectory stream with the corresponding body parts (see Fig.\ref{fig:overview}f). 

Before performing this association, we first remove erroneous part detections which can readily be identified as outliers. 
We find that a simple yet robust method is to use the depth maps from the multiple RGB+D sensors. At any time instant, a part can be considered as an outlier if it is {\em outside} of every surface in the dense point cloud. We simply test this by checking whether a part proposal is in front of the measured depth in any view, and mark it as erroneous if it is. 
This is a necessary but not sufficient condition because we test this from only the 10 available depth map views.
However, we find that this method works well in practice and is efficient to implement. After identifying these outliers, we remove and treat them as missing data. Then, we can assume that this filtered part trajectory only suffers from relatively small jitter and occasionally missing data. 

We associate a set of patch trajectories with a filtered part trajectory proposal if they move rigidly and the patch normal is a match. Intuitively, the part should be located inside the body surface, and, thus, a vector from the closest point on the part to the patch center should have a similar direction as the patch normal---their inner product should be positive. For a part trajectory proposal, we only consider patches for which the normal satisfies this criterion for the entire duration of the patch trajectory. As additional criteria, we compute a measure of rigidity between a patch trajectory and a part trajectory proposal. We define this as the difference between the minimum and maximum distance between them across all frames $t$ in which they overlap:
\begin{gather}
d(\mathbf{f}_i,\mathbf{\tilde P}_{uv}^k) = \max_t \, l\!\left(\mathbf{f}_i(t),\mathbf{\tilde P}_{uv}^k(t)\right)  - \min_t \, l\!\left(\mathbf{f}_i(t), \mathbf{\tilde P}_{uv}^k(t)\right),   \nonumber
\end{gather}
where $l( \cdot, \cdot )$ is the orthogonal distance between the patch center and the line segment of the body part, i.e.,
\begin{gather}
l\!\left(\mathbf{f}_i(t),\mathbf{\tilde P}_{uv}^k(t)\right) = \min_\alpha \| \alpha \mathbf{N}_{u}^{k_u}(t) {+}(1{-}\alpha)\mathbf{N}_{v}^{k_v}(t)  - \mathbf{f}_i(t)\|_2. \nonumber
\label{eq:segment_distance}
\end{gather}
Here, the set of time instants $t$ satisfies that both the patch trajectory and part trajectory streams are valid, and only patch trajectories $i$ for which $0{\leq}\alpha{\leq}1$ at some time $t$ are considered. Intuitively, this cost approximates how rigidly they move together over time, going to zero for completely rigid motion. Each part trajectory $\mathbf{\tilde{P}}^k_{uv}$ is then associated with a set of patch trajectories $\mathbf{F}^k_{uv}$, for which the rigidity cost is less than a threshold, i.e., $\mathbf{F}^k_{uv} = \{ \mathbf{f}_i : d(\mathbf{f}_i,\mathbf{\tilde P}_{uv}^k){\leq}10\textrm{cm}\}$. If a patch trajectory is selected by multiple body parts (e.g., a static scene as an extreme case), the trajectory is associated with the body part with minimum $\max_t \, l (\mathbf{f}_i(t),\mathbf{\tilde P}_{uv}^k(t) )$ distance. An example of this labeling is shown in Figure~\ref{fig:overview}f.

\subsection{Motion Refinement by Associated Patch Trajectories}
From the set of patch trajectories $\mathbf{F}_{uv}^k$ associated to the part trajectory proposal $\mathbf{\tilde{P}}_{uv}^k$, we can compute the rigid transform between subsequent time instances from $t$ to $t{+}1$, $T(t{+}1\left|\right.t) $, and, progressively, to any frame $t'$ by concatenating transformations between subsequent frames, so that $T(t'\left|\right.t)\mathbf{\tilde{P}}_{uv}^k(t)$ represents the propagated part from time $t$ to $t'$. Using the transformation it is possible to propagate a body part's position to other time instants. Our method uses the propagated parts to reduce jitter and fill in missing holes by averaging multiple part locations propagated from different time instances. For a target time $t$, we can produce multiple proposals for the same part, including the proposal from the first stage of our method and propagated parts using the estimated transformations, creating a set 
\begin{gather}
\{ T(t\left|\right.t{-}n) \mathbf{\tilde{P}}_{uv}^k(t{-}n),\cdots\!,\mathbf{\tilde{P}}_{uv}^k(t),\cdots\!,T(t\left|\right.t{+}n) \mathbf{\tilde{P}}_{uv}^k(t{+}n) \}.		\nonumber
\end{gather}
If there are elements in this set, we take the average. If the set is empty due to consistently severe occlusions, we determine that the part at time $t$ is still missing. In practice, we use $n{=}$1. This procedure can also be iterated multiple times (including patch trajectory re-association) to fill in missing parts that are further than $n$ frames from any part proposal. We iterate this refinement until no more missing parts can be filled. After refinement, a node connected to multiple body parts can have different locations corresponding to each of the averaged parts, and we simply take the average to determine the final node locations. It should be noted that our method is different from temporal smoothing (e.g., ~\cite{Elhayek-16}). Instead, we use an actual measurement of 3D motion rather than impose a motion prior, which prevents over-smoothing even after several iterations.

\section{Results}

We quantitatively and qualitatively evaluate our method on various sequences captured in the Panoptic Studio. Our dataset includes diverse social games performed by multiple people. In the quantitative evaluation, we empirically show how the large number of views solves the challenging interaction capture problem; we compare performance using varying number of cameras on the scenes with different number of people. In the qualitative evaluation, we demonstrate the ``model-free" advantage of our method by showing compelling motion reconstruction results on subjects of diverse appearance, body shapes, and body sizes.

\begin{figure}[t]
	\centering
	\subfigure[Mafia]{\includegraphics[width=0.24\textwidth]{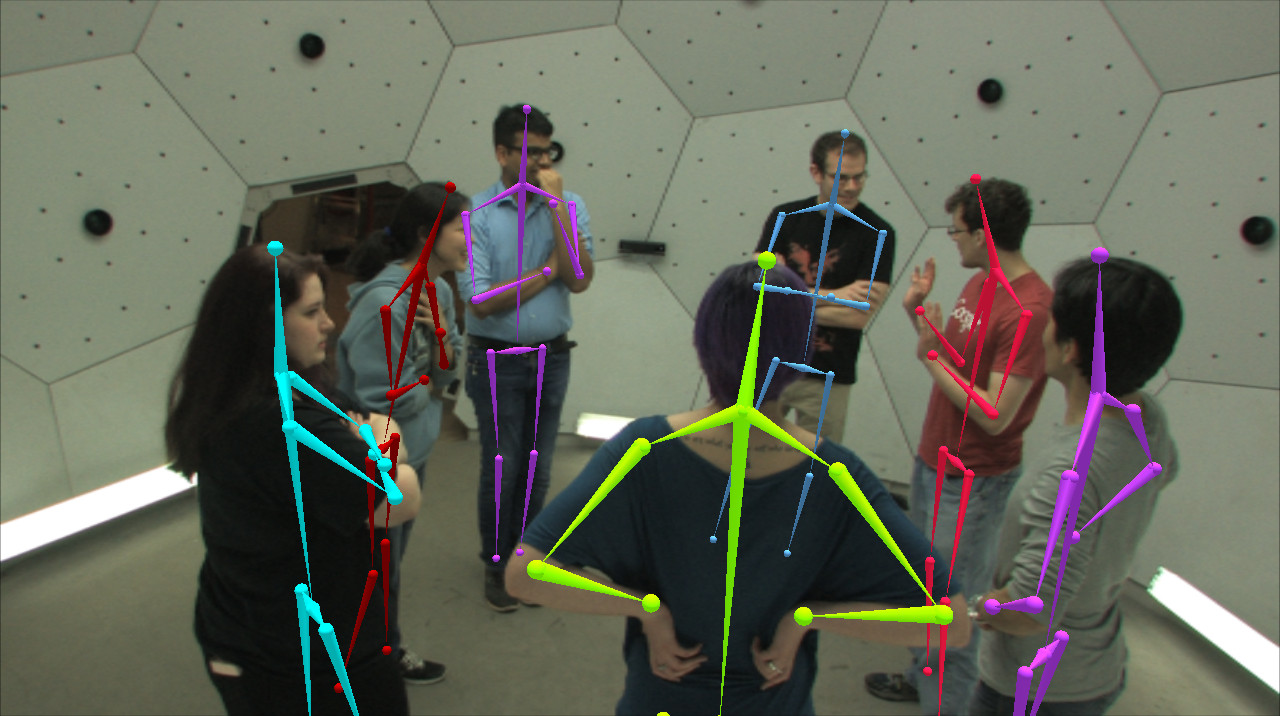}} 
	\subfigure[Ultimatum]{\includegraphics[width=0.24\textwidth]{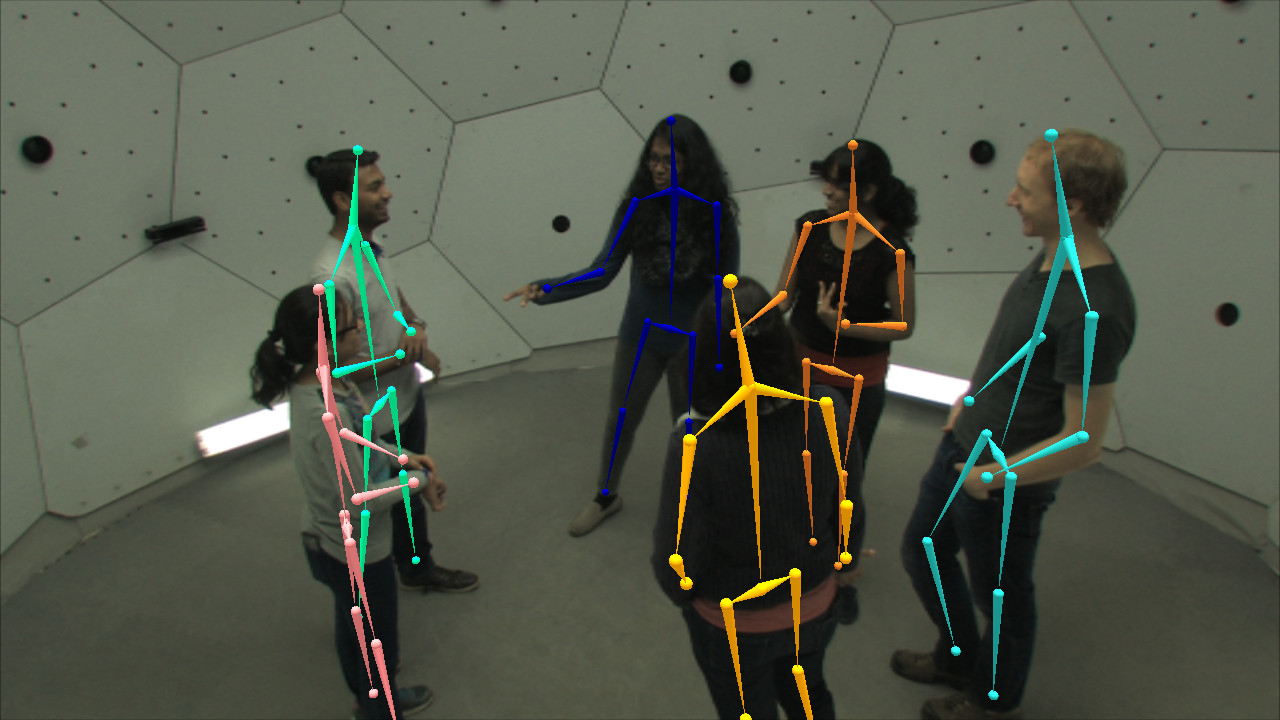}} 
	\subfigure[Haggling]{\includegraphics[width=0.24\textwidth]{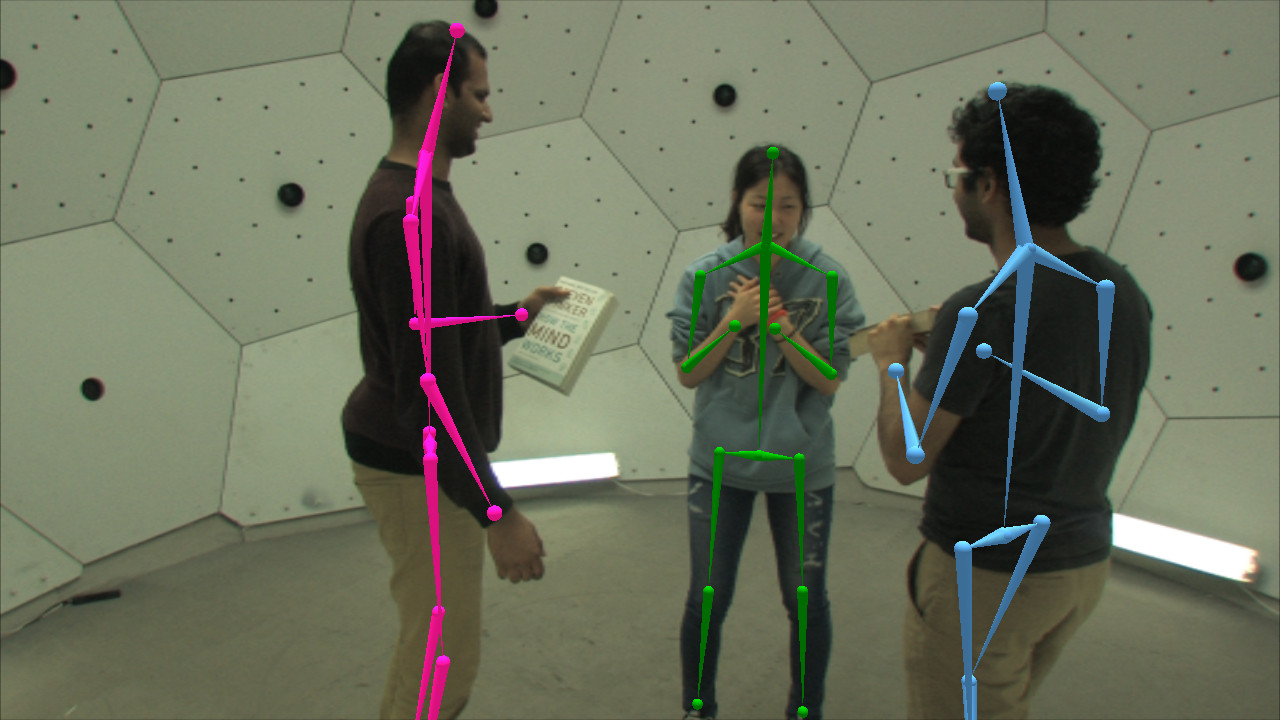}}   
	\subfigure[007-Bang]{\includegraphics[width=0.24\textwidth]{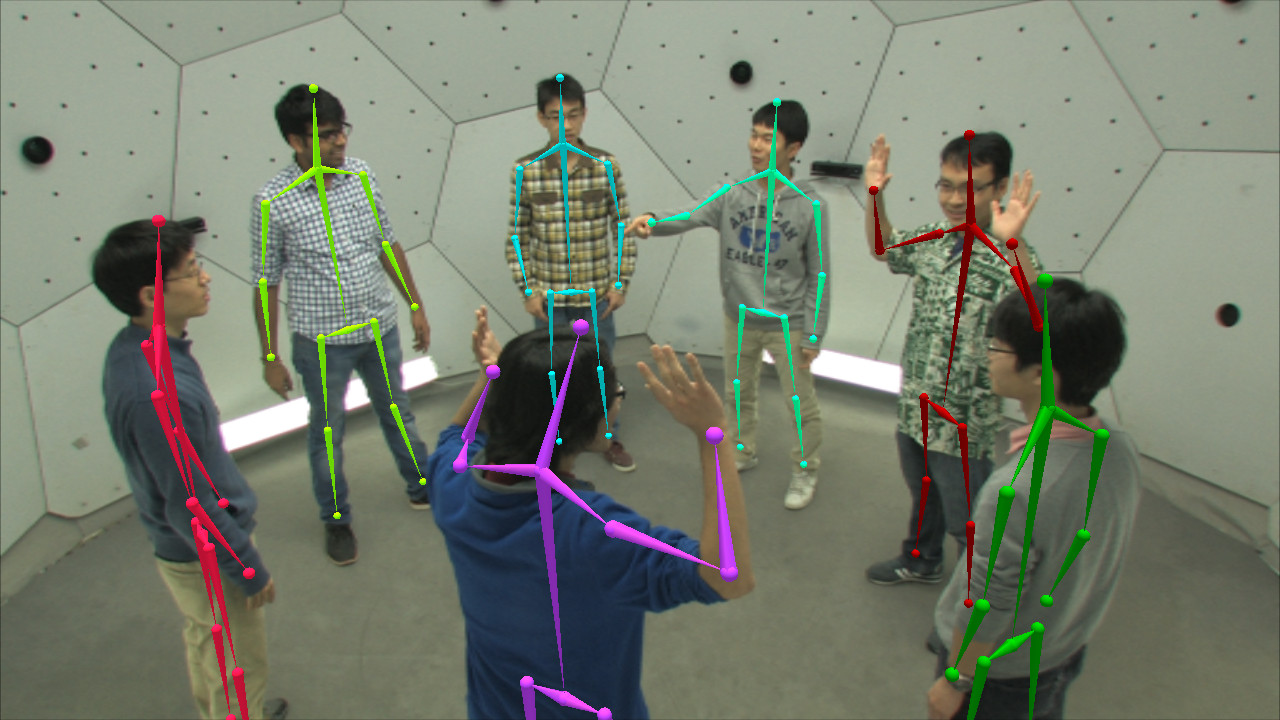}}   
	\caption{Example scenes of social game sequences. The reconstructed 3D skeletons from the 480 VGA views are projected on novel HD views.} 
	\label{fig:socialGames}
\end{figure}

\subsection{Dataset and Capture Procedures}
We captured a group of people engaged in social interactions using the Panoptic Studio\footnote{Some sequences were captured with fewer than the full set of cameras due to hardware failures during capture.}. To evoke natural interactions, we involved participants in various games: \emph{Ultimatum}, \emph{Mafia}, \emph{Haggling}, and \emph{007-Bang Game}. The first two games are used in experimental economics and psychology to study conflict and cooperation, and the latter two games also induce a variety of rich non-verbal signals in participants. Example scenes of each game are shown in Figure~\ref{fig:socialGames}. Refer to the supplementary material for descriptions of the games and capture procedures. In our captures, subjects were informed of the rules of the game but were otherwise not instructed about how to behave, nor was their clothing or appearance controlled. They were also not initially aware of our research goals to avoid potential biases in their gestures\footnote{The majority of the sequences are captured with people randomly recruited from a university campus; some sequences were captured for testing purposes and feature researchers with knowledge of the project. Those sequences are marked in our dataset website.}. The scenes in our dataset contain various natural motions which may commonly occur in the interactions of daily life, as shown in Figure~\ref{fig:iconicPoses} and \ref{fig:qualitativeSocial}. 

To additionally demonstrate the performance of our system and methods, we capture other challenging sequences, including a group of 8 seated people participating in a discussion (\emph{meeting} sequence), a mother and a toddler at play (\emph{toddler} sequence), musical performances with severe occlusions due to the instruments (\emph{drummer} and \emph{cellist}), and a sequence featuring various fast motions and challenging postures (\emph{dancer}). 

In aggregate, the dataset contains about 198 minutes ($\sim$297K frames) of videos, for a total of about 154 million images. Our dataset is summarized in the supplementary material. 

The main distinguishing features of this collection compared to previous markerless motion capture datasets are: (1) natural interactions in the scenes showing rich and subtle non-verbal cues, (2) social groups of up to 8 interacting people, and (3) coverage by a large number of views (up to 521). We make all the data available on our website, including all synchronized camera feeds, calibration, 3D pose reconstruction results, and 3D trajectory streams: \url{https://domedb.perception.cs.cmu.edu}. 

\begin{figure*}[t]
	\centering
	\captionsetup{position=top}
	\subfigure[Two people]{\includegraphics[trim=40 180 60 170,clip,width=0.23\textwidth]{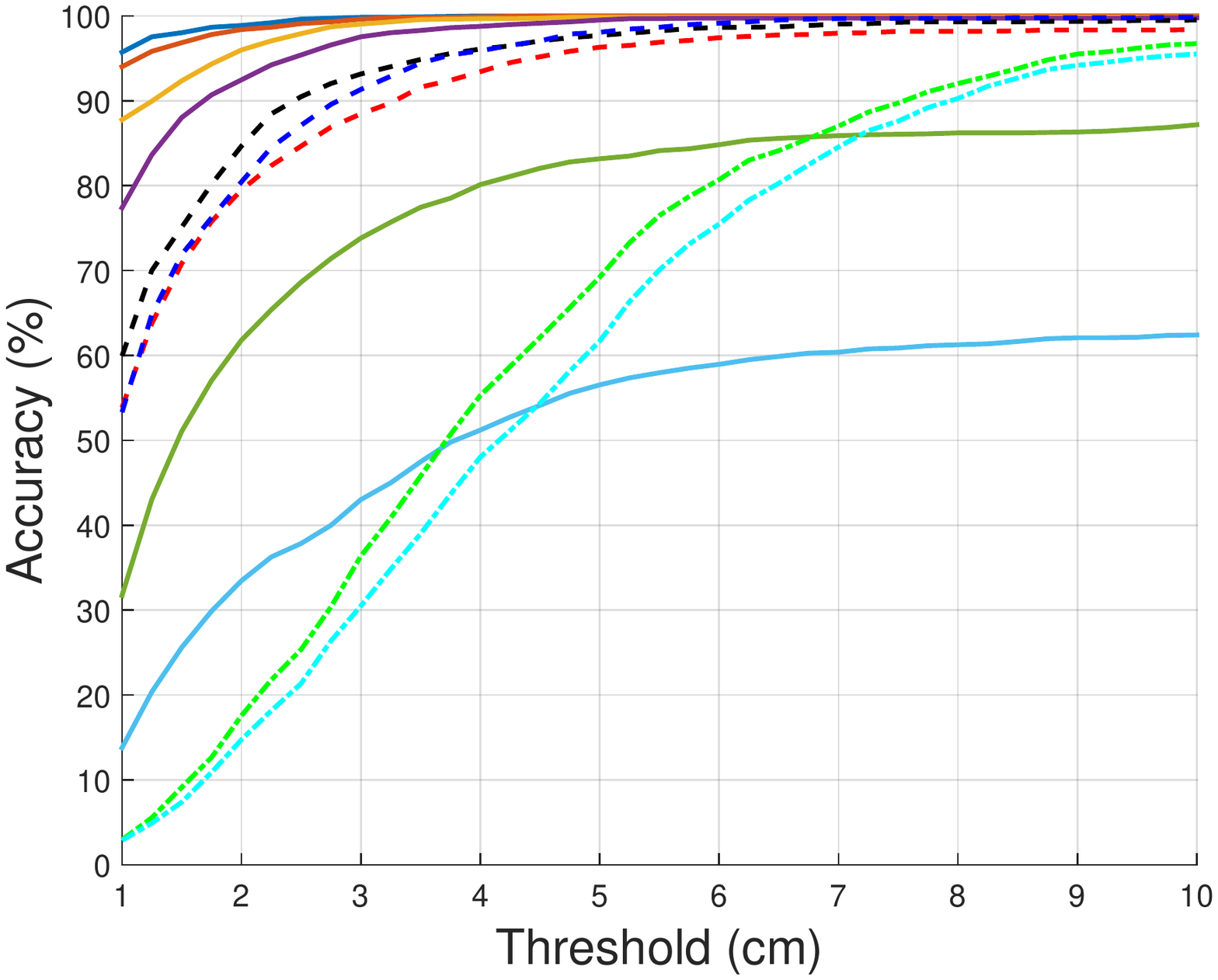}
		\llap
		{\shortstack{%
				\includegraphics[width=0.1\textwidth]{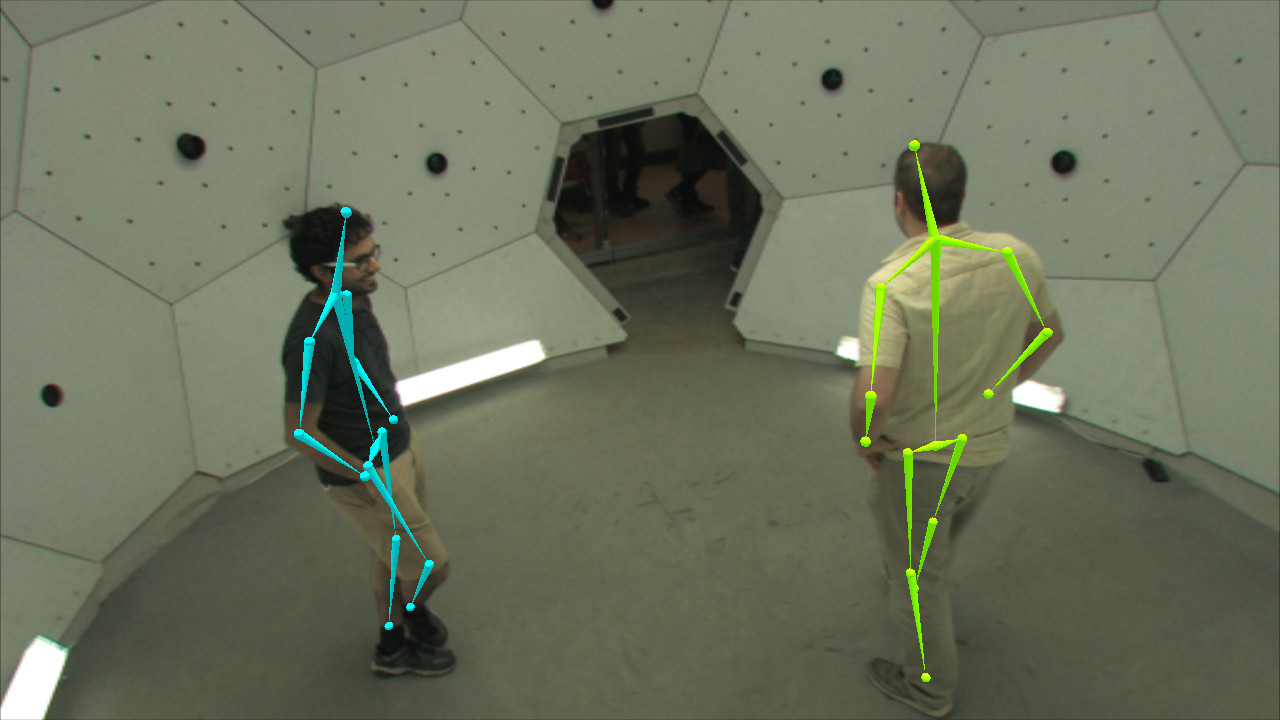}\\
				\rule{0ex}{3ex}	
			}
			\rule{1ex}{0ex}	
		}	
	}
	\subfigure[Three people]{\includegraphics[trim=40 180 60 170,clip,width=0.23\textwidth]{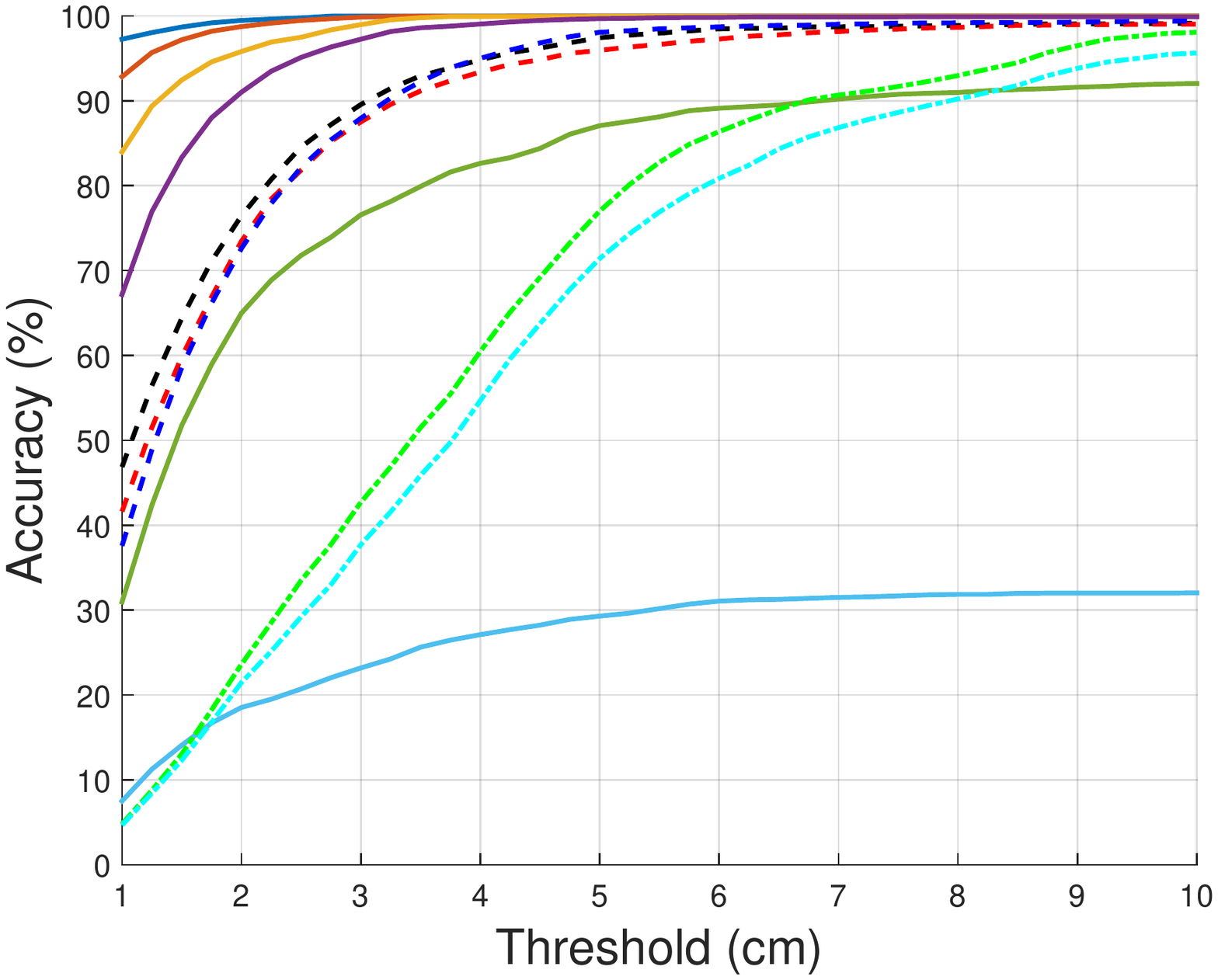}
		\llap
		{\shortstack{%
				\includegraphics[width=0.1\textwidth]{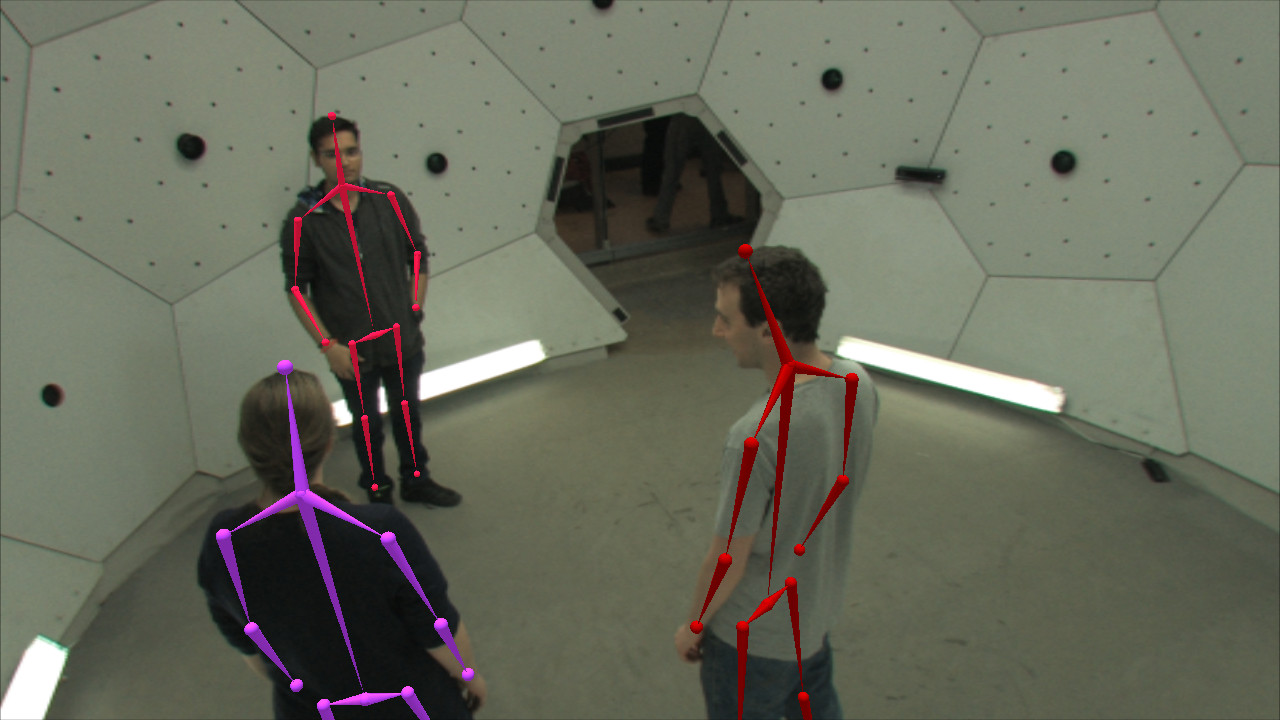}\\
				\rule{0ex}{3ex}
			}
			\rule{1ex}{0ex}
		}	
	}
	\subfigure[Five people]{\includegraphics[trim=40 180 60 170,clip,width=0.23\textwidth]{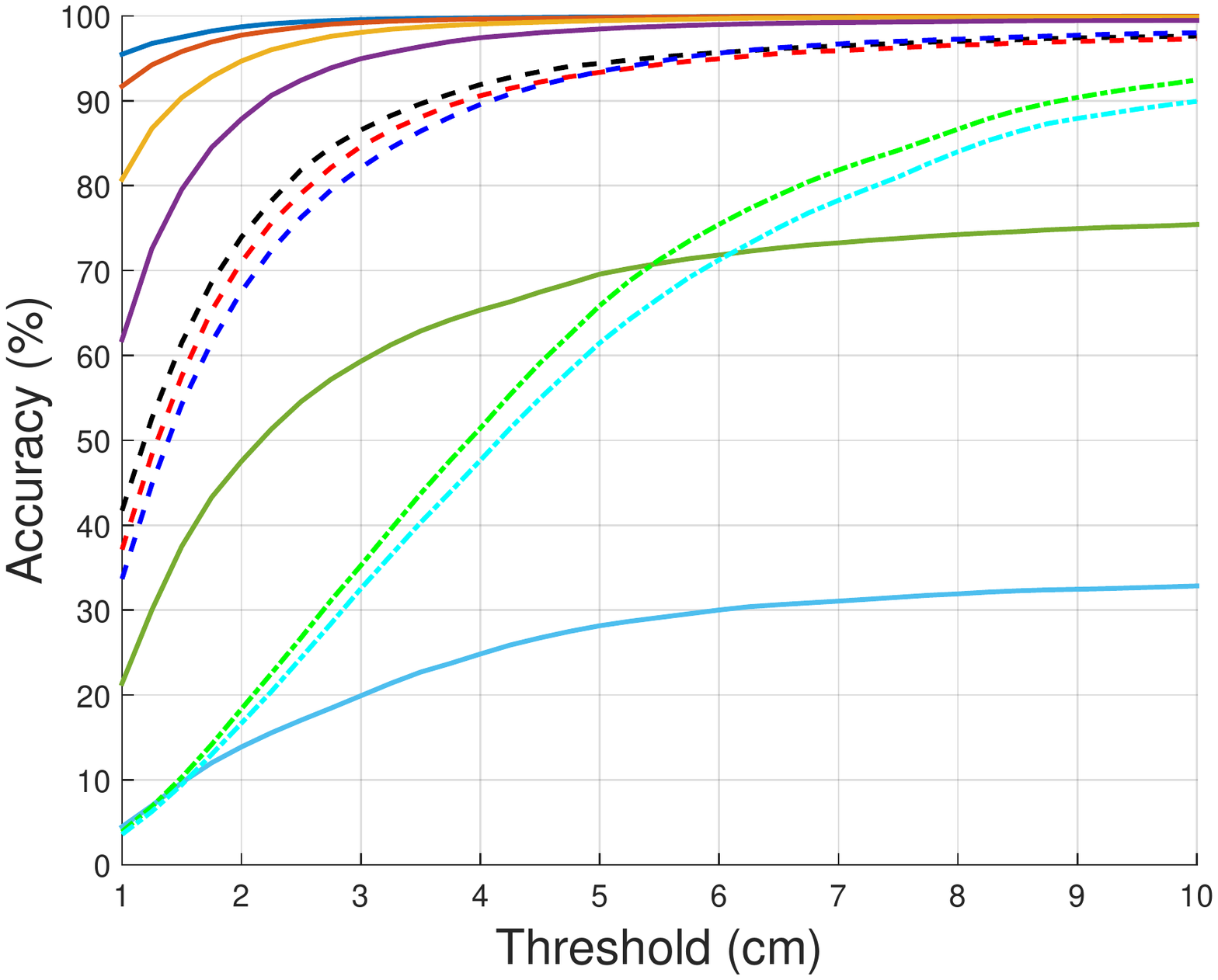}
		\llap
		{\shortstack{%
				\includegraphics[width=0.1\textwidth]{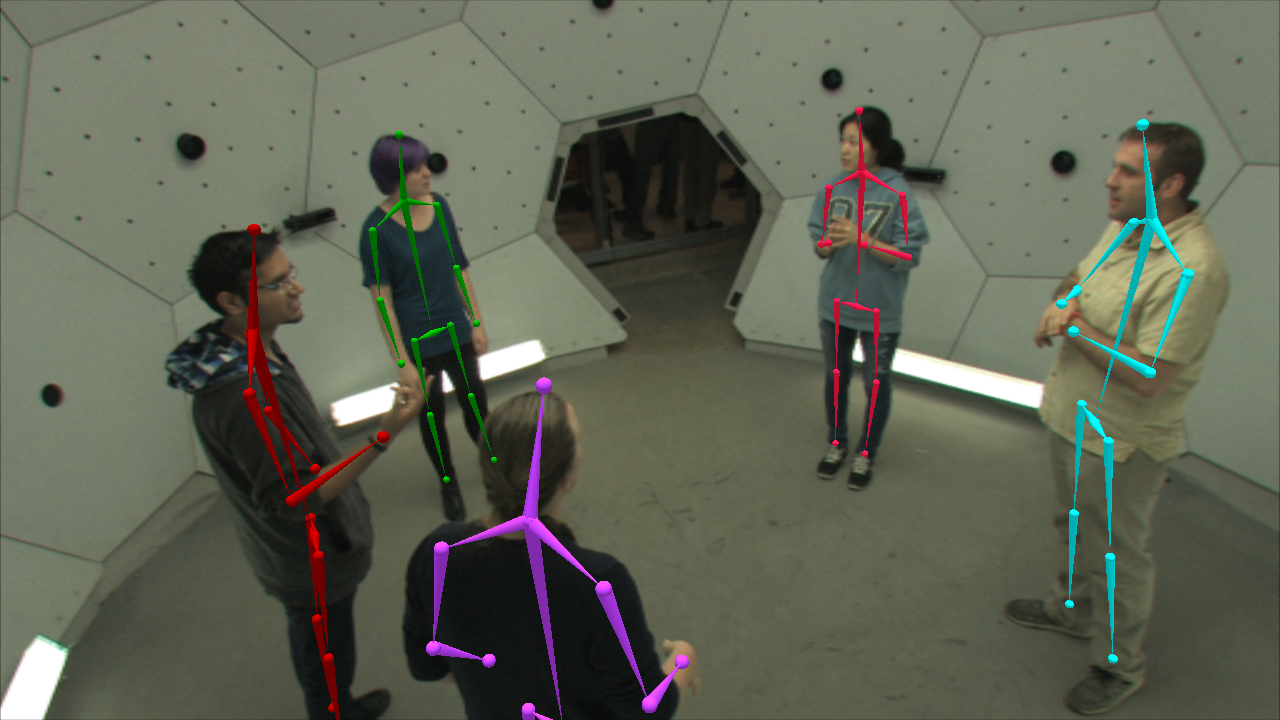}\\
				\rule{0ex}{3ex}
			}
			\rule{1ex}{0ex}
		}	
	}
	\subfigure[Seven people]{\includegraphics[trim=40 180 60 170,clip,width=0.23\textwidth]{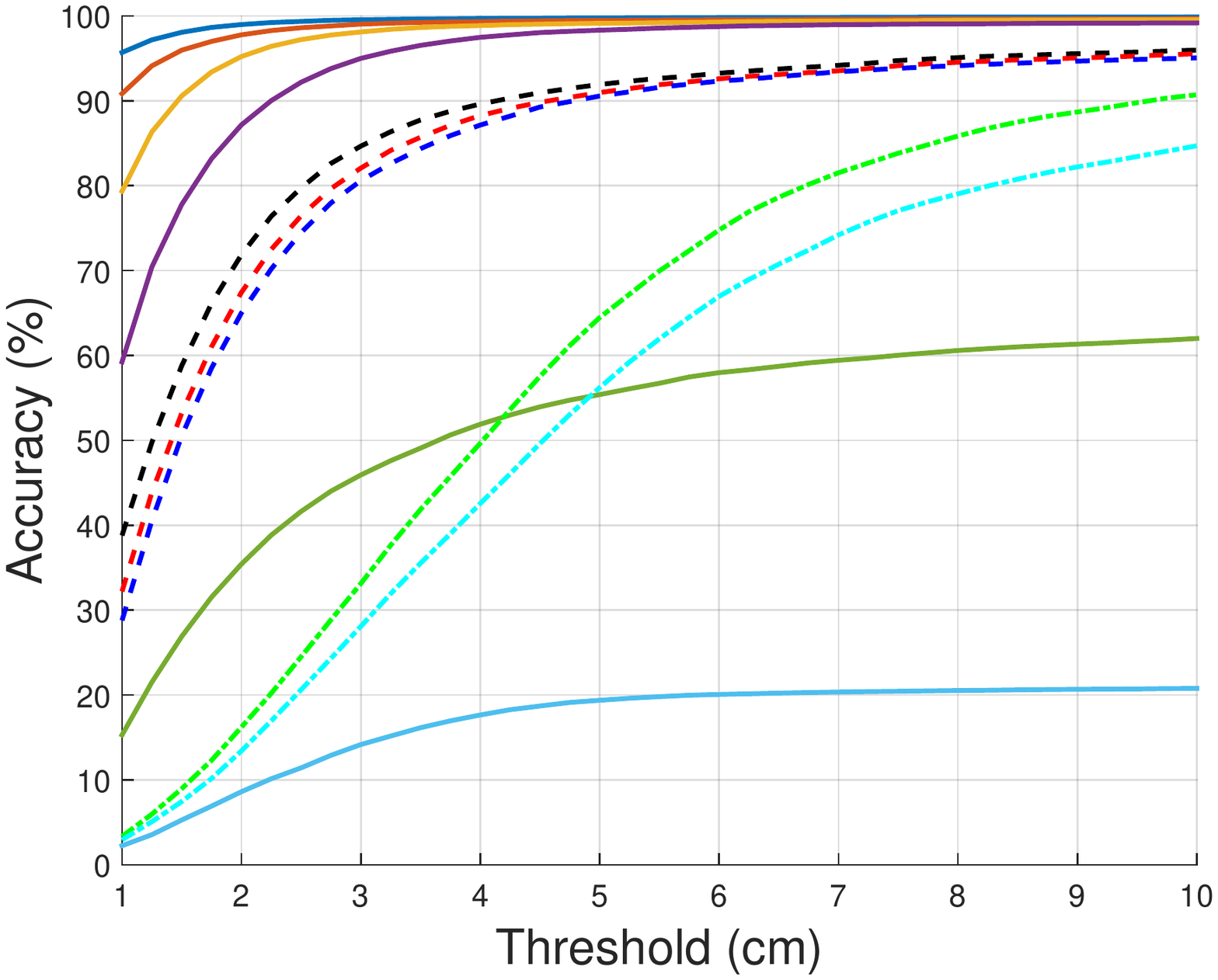}
		\llap
		{\shortstack{%
				\includegraphics[width=0.1\textwidth]{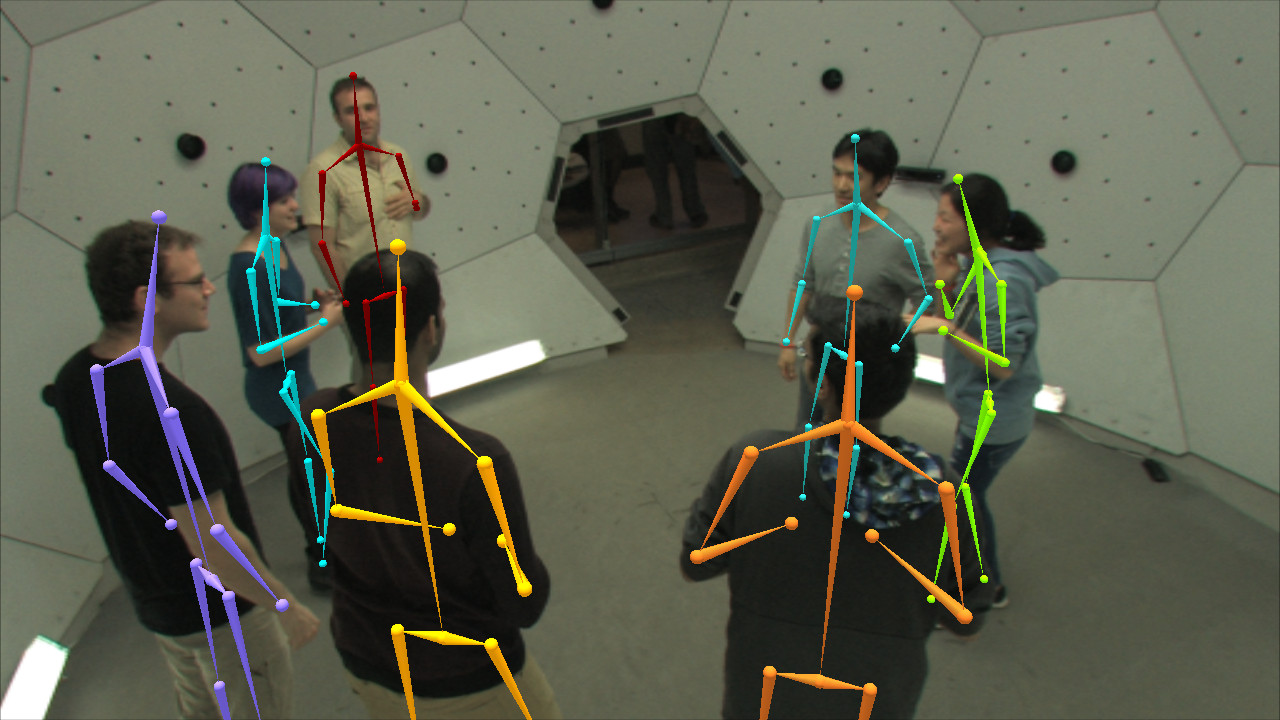}\\
				\rule{0ex}{3ex}
			}
			\rule{1ex}{0ex}
		}	
	}
	\subfigure{\includegraphics[width=0.9\textwidth]{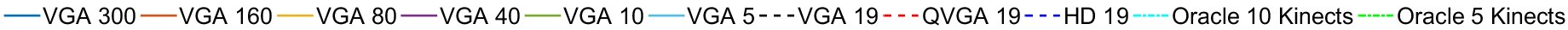}} 
	\caption{Performance evaluation using Probability of Correct Keypoint (PCK) metric for varying number and type of cameras on \emph{160422 ultimatum1}. We use the result of 480 VGA cameras after manually excluding outliers as ground truth. The X-axis of each graph represents thresholds, and the Y-axis represents accuracy by the thresholds. Each graph is generated for scenes with a different number of people. The results demonstrate that more views (rather than higher resolution) are beneficial to improve accuracy, and the distinction is more noticeable if the scene contains more people.} 
	\label{fig:quant1}
\end{figure*}

%
%

\begin{table} [t]	
	\centering
	\caption{Processing time for one minute of data.}\label{Table:questionaire}
	\begin{tabular}{l | l|  r}
		\hline
		&  Procedure & Time  \tabularnewline
		\hline
		\multirow{4}{*}{Stage 1} & (\ref{subsection:nodeProposals}) 2D pose detection (1 GPU) &  40 h  \\
		
		&(\ref{subsection:nodeProposals}-\ref{subsection:partProposals})  Node and part proposal recon. (1 GPU) &  4 h  \\
		
		&(\ref{subsection:dynamicProgamming}) Skeletal proposal reconstruction by DP  &  3 m  \\
		
		&(\ref{subsection:dynamicProgamming}) Skeletal proposal optimization  &  11 m  \\
		\hline
		\multirow{2}{*}{Stage 2} & (\ref{subsection:patchTrajectoryRecon}) Trajectory stream recon. (400 CPU cores)  &  35 h  \\ 
		& (\ref{subsection:trajectoryAssociation}) Trajectory association and refinement &  5 m  \\
		\hline
	\end{tabular} 
	\label{table:processingTime}
\end{table}

\subsection{Processing Time}
The time to process one minute of data (1500 frames) of 480 VGA views is summarized in Table~\ref{table:processingTime}. We use different computing devices for procedures.  A machine with Intel i7 3.4GHz processor and 32GB RAM is used for general processing, a GTX Titan X is used for GPU computation, and a cluster server with 400 CPU cores (2.2GHz per processor) is used for trajectory stream reconstruction. 

In the first stage, most of the time is spent in running the 2D body pose detector. The detector runs at about 5 frames per second on a single GPU, but due to the large number of views (720K images per minute),  processing a minute of video takes about 40 hours. In practice, we use multiple GPUs to process multiple images in parallel. In the second stage, the main computational bottleneck is the trajectory stream generation. Although they are tracked in parallel, the running time is long due to the large number of patches at each time. In our experiments, on average 15K patches are tracked per person. 

\begin{table} [t]	
	\centering
	\caption{Quantitative evaluation of the accuracy of our method on the \emph{160422 ultimatum1} sequence. }\label{Table:otherDataset}
	\begin{tabular}{c|c|c|c|c}
		\hline 
		{Skel. \#} & {Node \#} & {Outlier Node \#} & {Node Acc.}  & {Skel. Acc.} \tabularnewline
		\hline 
		81,829 & 1,227,435 & 8700 & 99.29\% & 93.55\% \tabularnewline
		\hline 
	\end{tabular} 
	\label{table:quant_stat}
\end{table}

\subsection{Performance Analysis of The Panoptic Studio}

We quantitatively evaluate the performance of our method for the \emph{160226 ultimatum1} sequence by varying the number and type of cameras. We choose the ultimatum sequence because it captures varying number of people (from two to seven people) in each time period, which is suitable to study the relation between scene complexity and the number of cameras needed to reach a desired performance. In this experiment, we only evaluate the first stage of our method.  

\textbf{Performance using all VGA cameras}: We first quantify the performance of our system when all 480 VGA cameras are used. Due to the absence of ground truth data, we manually annotate the correctness of the reconstructed 3D skeletons by verifying their projections in multiple 2D views. We labeled a 3D joint node as an outlier if the node is projected outside of the corresponding limb or far from the presumable target joint in multiple 2D views. We exclude the period where people come in and out of the system, since at the moment body parts lie on the edge of our system's working volume. The result of the quantitative evaluation for the 15 min. of sequence is summarized in Table~\ref{table:quant_stat}. There are 12 sessions in the sequence, and 61 temporally associated skeletal structures are reconstructed. Among about 1.2 million body joints, about 8.7K nodes are determined as outliers or missed (rejected by thresholds of our system), showing 99.29\% accuracy in node reconstruction. And, 93.55\% out of about 82K 3D skeletons are correctly reconstructed without any incorrect joints. The majority of the failures are caused by insufficient visibility of the target part. An example is the pose holding hands behind one's back near the wall of the system as shown in Figure \ref{fig:failures} (left). Although the hands are visible from few cameras, they are too close to be detected by the pose detector. Interestingly, our method still reconstructs the hands using the ``guessed" 2D locations from 2D pose detector in frontal views, although the accuracy is limited as shown in Figure \ref{fig:failures}.

\textbf{Comparison with varying number of cameras}: To evaluate the impact of the number of views, we perform our method using varying number of cameras. The cameras are uniformly sampled (except the 19 VGA camera case explained later); i.e., we sample the next camera as the one furthest from all the already sampled cameras, and, thus, the selected cameras are always a subset of the set of the larger number of cameras. To quantify the results, we treat the result with 480 VGA cameras as ground truth after excluding the manually annotated outliers. For evaluation, we only use every tenth frame to reduce computation time. As an evaluation metric, we use the PCK (Probability of Correct Keypoint) metric, which is commonly used to evaluate 2D pose detectors~\cite{Andriluka-14}. Here, we use 3D distance in physical scale (cm) obtained from calibration data for the threshold of PCK, in contrast to the 2D ratio of torso/head as in 2D pose detection cases~\cite{Andriluka-14}. Figure~\ref{fig:quant1} shows the PCK accuracy by varying the camera number on the scenes with different number of people. In all the results, we find that using a larger number of views is beneficial. If the scene is simpler (e.g., the case with two or three people), we observe that the results with a smaller number of cameras, e.g., 160 cameras, show a similar performance with 480 cameras. However, if the scene becomes more complicated, e.g., seven people, we see clearer gaps according to the camera numbers. This results can be meaningfully used to design a multiple camera system to determine the required number of cameras given a desired group size. For example, assuming that the target scenes have about five people, we forecast that a system with 80 cameras can reach about 94\% of accuracy with a 2cm threshold.

\begin{figure}[t]
	\centering
	\captionsetup{position=top}
	\includegraphics[trim=40 340 60 340,clip,width=\columnwidth]{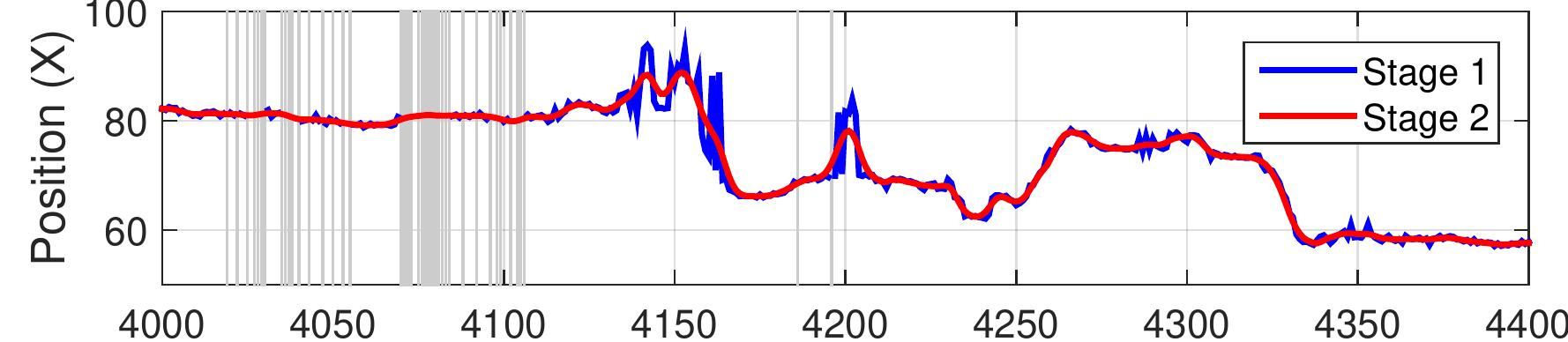}   \\
	\caption{Refinement result by the Stage 2 of our method on the \emph{160226 mafia2} sequence. The most erroneous node is selected. The graph represents the X coordinate of the node across frames after the Stage 1 method (blue) and Stage 2 method (red). The gray regions represent the frames where the part is missing in the stage 1 output. They are recovered via the temporal propagation in the result of Stage 2. } 
	\label{fig:quant_stage2}
\end{figure}


\textbf{Comparison with varying resolutions}: As an additional evaluation, we perform a similar experiment for different camera resolutions using the multiple HD cameras installed in our system. Among 31 HD cameras, we use 19 HD cameras installed on the same panels with VGA cameras\footnote{We have 20 HD cameras installed on the same panels with VGAs, but we missed 1 HD camera due to the hardware failure during the capture.}.  To compose similar viewpoints, we choose the closest VGA cameras from the selected 19 HDs. Additionally, we generate 19 QVGA inputs (320 $\times$ 240 resolution)  by resizing the selected VGA videos. Because the HD cameras are not perfectly synchronized with VGAs, we interpolate the result from HDs into the VGA time domain using the hardware sync data. The performance of a same number of HDs, VGAs, and QVGAs is shown as dashed lines in Figure~\ref{fig:quant1}. The result shows that the performance differences among them is marginal, although HD views have about 7 time more pixels than VGAs and about 27 times more pixels than QVGAs. The result demonstrates that the pose reconstruction performance of our method is marginally affected by the resolution changes compared to the changes by the number of views. Note that the integral of number of pixels in the 19 HD views are equivalent to about 128 VGA views, and the result clearly shows that it is more advantageous to have more unique camera views rather than having higher resolutions, given a fixed pixel budget. The main reason underlying this finding is that dealing with occlusions is more crucial in interaction capture scenarios, and, in particular, higher resolution is not beneficial in our method, since 2D joint localization accuracy is still limited by the 2D pose detector.

\textbf{Comparison to multiple Kinects}: We also compare our results with the result of multiple Kinects. Since Kinect with its accompanying SDK is one of the most commonly used sensors for markerless motion capture in various communities, using multiple Kinects can be considered as an option to handle severe occlusions for interaction capture. However, how to fuse multiple Kinect cues is not straightforward, and, thus, we naively fuse them as follows. We first generate 3D skeletal proposals from all individual Kinects, and simply find the best candidate closest to our ground truth data in Euclidean space, assuming that an Oracle chooses the best one given the GT data. This can be considered an upper bound of a naive multiple Kinects method. Since the keypoint locations of the Kinects are not identical to the skeletons of our method, for a fair comparison, we adjust the Kinect skeletons by finding an offset vector from each Kinect node toward our node of GT's skeleton in a person-centric coordinate system. As shown in Figure~\ref{fig:quant1}, the results of the \emph{Oracle} Kinects is limited, showing less than 80\% accuracy at a 5cm threshold. 

\subsection{Refinement by Trajectory Stream}

We compare the performance improvement of our refinement method (the second stage) over the output of the first stage. We choose a  challenging scene in \emph{160226 mafia2} sequence where the first stage of our method shows failures due to the erroneous 2D pose detection results. To see the performance change, we plot the X coordinate of the most erroneous node (right wrist of a subject) as shown in Figure~\ref{fig:quant_stage2}. The frames denoted as gray regions are the time when the nodes are missed due to the consistent 2D pose detector failures. It is shown that our refinement method can recover the missing parts and also noticeably reduce the motion jitter for the unstable frames. Note that our refinement method is not just smoothing but based on the temporal transformation measured by a dense trajectory stream. Thus, it does not suffer from over-smoothing, even after several iterations. 
%
%

\subsection{Comparison with The Method of Joo et al.~\cite{Joo-15}}
We compare the presented method to the method introduced in \cite{Joo-15}. In \cite{Joo-15}, due to the relatively unreliable 2D pose detection cue~\cite{Yang-2012}, the motion cues from trajectory stream play a core role to reconstruct valid parts. The method, however, tends to fail in regions where the trajectory stream is unavailable (e.g., the texture-less dark body parts). The method presented in this paper is composed of two sequential stages using an advanced 2D pose detector~\cite{Wei-2016}, and Stage 1 is still applicable in the region where trajectory stream is unavailable. Table~\ref{Table:iccvComparison} shows the comparison between two methods on the sequence \emph{150129 007-Bang} introduced in \cite{Joo-15}, where the accuracy is computed by manually annotating outliers. The major failures of \cite{Joo-15} occur on the texture-less leg parts, or fast motion with motion blur when the trajectory stream is sparse and inaccurate. Refer to the supplementary video for the qualitative comparison. 

\begin{table} [t]	
	\centering
	\caption{Quantitative comparison to \cite{Joo-15} on the \emph{150129 007Bang} sequence. }
	
	\begin{tabular}{c|c|c|c|c|c}
		\hline 
		{Methods} & {Frames} & {Joints} & {Outliers} & {Missed} & {Accuracy}\tabularnewline
		\hline 
		Ours & 300 & 22,500 & 1 & 0 & 99.99\% \tabularnewline
		\hline 
		\cite{Joo-15} & 300 & 22,500 & 1871 & 2248 & 80.80\% \tabularnewline
		\hline 
	\end{tabular} 
	\label{Table:iccvComparison}
\end{table}

\subsection{Qualitative Evaluation}
\begin{figure*}[t]
	\centering
	\includegraphics[trim=0 0 0 0,clip,width=\linewidth]{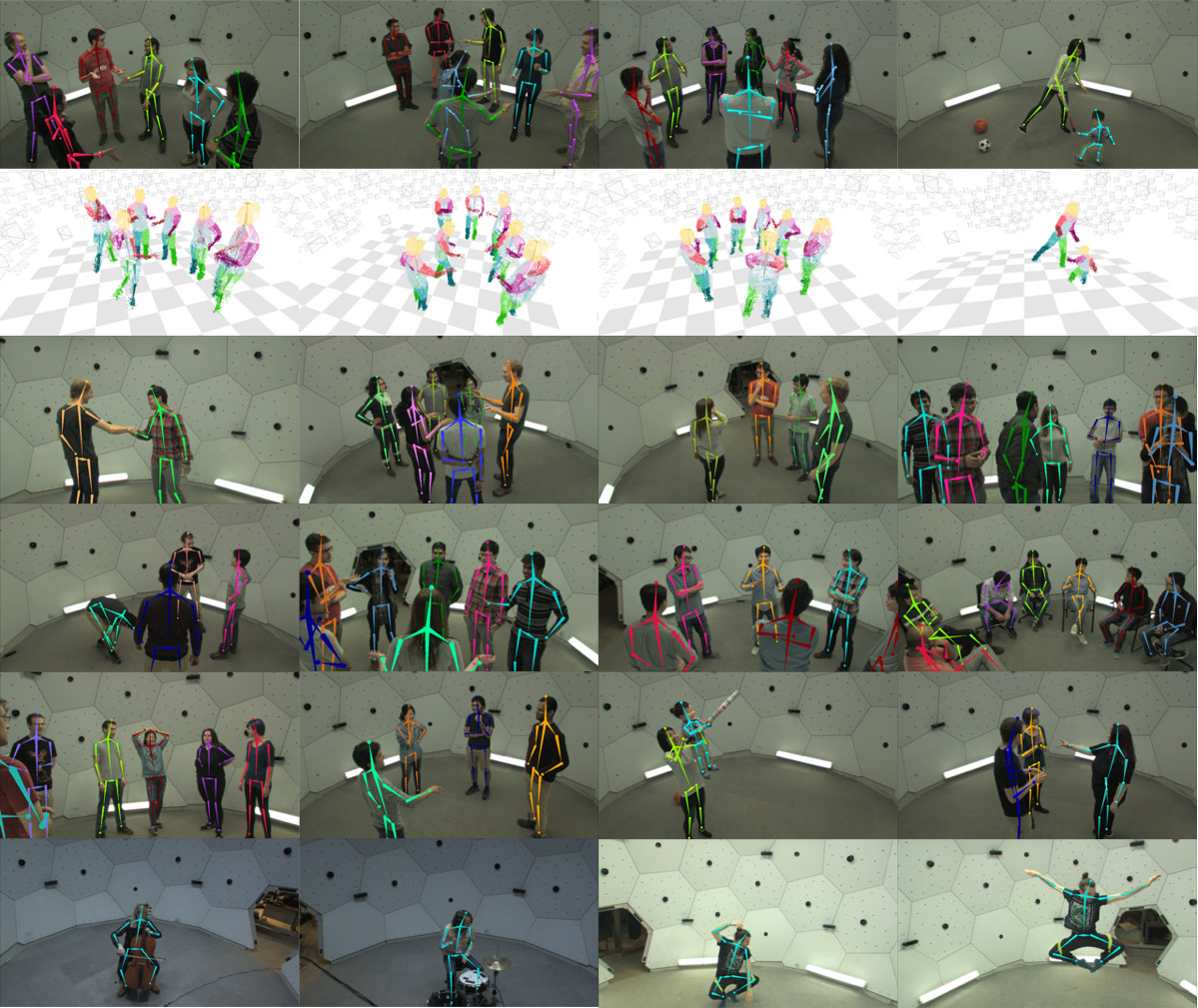}
	\caption{ Our method is performed on various scenes including social interactions of multiple people. (Row 1) Reprojected skeletons on novel HD views; (Row 2) Rendered 3D skeletons in novel 3D views with node trails over time; (Row 3) Labeled 3D trajectories representing articulated non-rigid body parts with same color representations as in Figure \ref{fig:overview}; (Row 4-7) Reprojected skeletons on novel HD views of various scenes.}
	\label{fig:qualitativeSocial}
\end{figure*}

We apply our method, producing about 3 hours of interaction capture results. Due to the computation time, the second stage of our method is applied on a subset of the dataset; yet the first stage of our method is applied on all the sequences\footnote{Results will be updated in our website, as they are processed.}. Example results are shown in Figure~\ref{fig:qualitativeSocial}. Our result is fully automatic---given video streams and calibration data, our method generates temporally associated 3D skeletons (and labeled patch trajectory stream of each body part if the second stage is applied) for each individual without any human supervision. Refer to the supplementary videos and the live 3D viewer on our website. 

\textbf{Group interaction capture}: Our method produces motion capture results on various social game scenarios performed by multiple people (up to 8 people). The number of subjects in the scenes is automatically determined by our method, and allowed to vary during the capture. The reconstructed results contain motions that frequently occur during communication, such as crossed-arms-on-chest, resting-chin-on-hand, mouth-guard, hands-on-back, hands-on-waist, and so on. In spite of their importance as non-verbal signals transmitting a variety of messages, such motions get little attention by prior work. In particular, severe topological changes and self-occlusions make it hard to apply 3D template-based motion capture approaches. Our method reconstructs the motion of such challenging scenes by fusing 2D pose detection cues and motion cues using a larger number of views, and demonstrates a compelling performance for social interaction capture. 

\textbf{Robustness to appearance, body sizes, and topological changes}: 
Our results demonstrate robustness to subjects of diverse appearance, body types, and sizes. As mentioned, subjects' clothing is not controlled, and the captured sequences contain people with various clothing such as black pants, thick padding jumpers, hoodies, short pants, scarfs, and so on. During the discussions, they also unconsciously adjust their clothing, for example by rolling up sleeves or relocating scarfs. The height of subjects varies from a two-year old toddler to adults more than 190 cm tall. The ``model-free" nature of our method enables us to reconstruct their motions without changing any parameter. It demonstrates a major advantage of our system for social behavior studies in that it can be easily applied for captures at scale, without any laborious template generation or initial alignment step. Especially, the toddler scene is challenging to  ``model-heavy" approaches, since instructing young children to be stationary to generate their template models (e.g., laser scanning) may not be practical. 


\textbf{Other interesting scenes}:
We also demonstrate the performance of our method on other atypical motion capture scenarios including musical performances (\emph{drummer} and \emph{cellist}) and \emph{dancer} sequences. Motion capture for musical performance is a good application for markerless motion capture, because markers may interfere with their functional movements during the capture. Although the scenes are challenging due to the severe occlusions by musical instruments, our method shows good performance in reconstructing the performer's subtle motions (e.g., the vibrato motion in the \emph{cellist} sequences).

On the other hand, the \emph{dancer} sequences contain fast motion and unusual poses. Due to failures in reconstructing the trajectory stream for the extremely fast movement compared to our relatively low frame rate cameras (25 Hz), we only apply the first stage of our method. Separating reconstruction (Stage 1) from temporal refinement (Stage 2) is advantageous in this case, because the first stage, based on per-frame reconstruction, is not affected by motion magnitude and free from error accumulation. We can optionally apply temporal refinement (Stage 2) based on the quality of trajectory stream to further refine the results. We find that, however, in a few extremely unusual poses our method becomes unstable due to consistent 2D pose detection failures, which will be discussed in subsection~\ref{subsection:Limitation}.

\begin{figure}[t]
	\centering
	\includegraphics[width=\columnwidth]{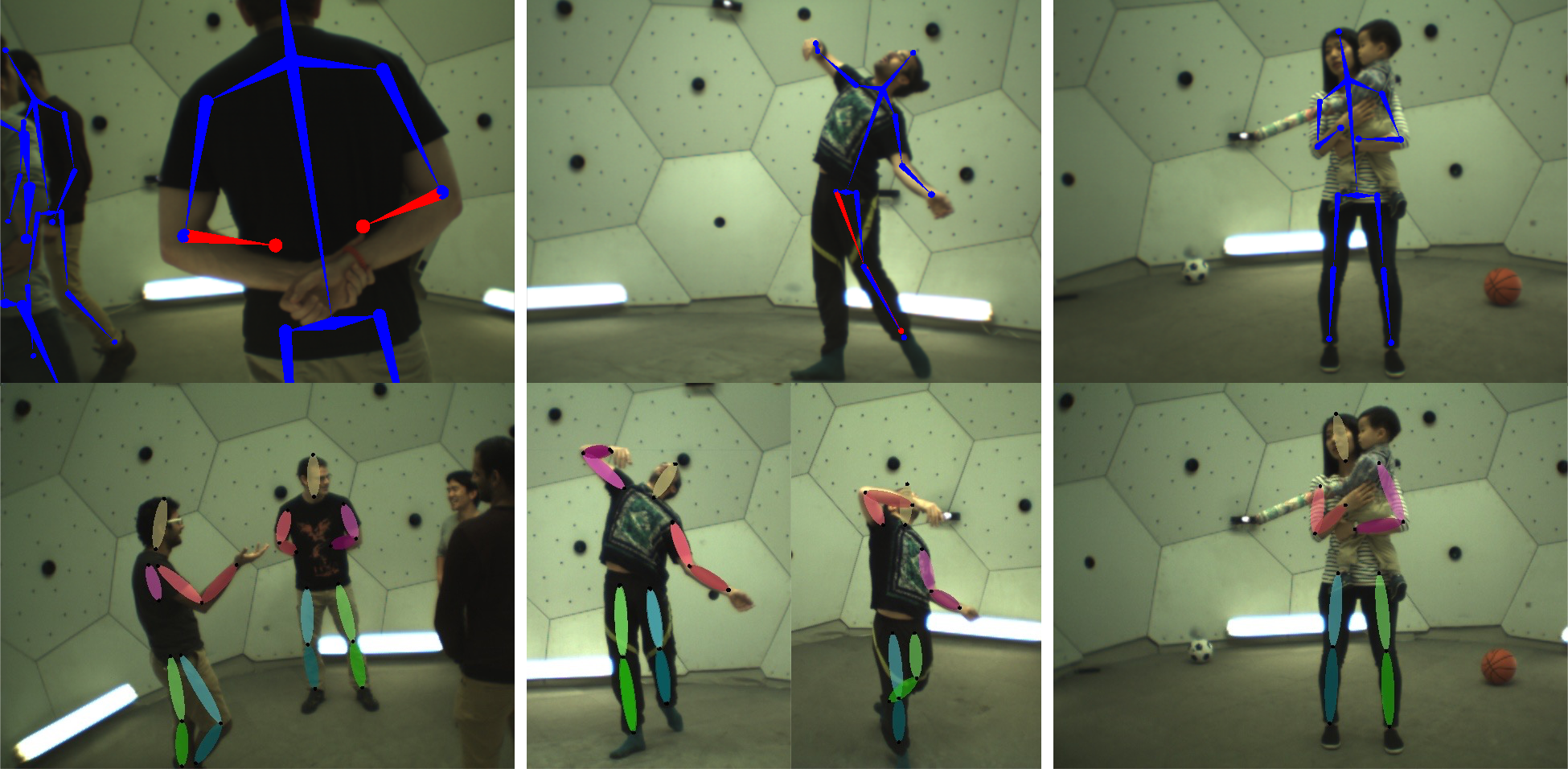}
	\caption{Example failure cases. For each column, the first row shows the projection of reconstructed 3D skeletons on a view where the red colored parts are manually annotated outliers. The second row shows the 2D pose detection results. (Left) The hands are severely occluded and only visible from few cameras where they are too close to be detected by 2D pose detector. (Center) The left/right legs are confused in performing 2D pose detection, which causes failures in our 3D inference. (Right) The toddler is not detected by the pose detector, since he is severely occluded. }\label{fig:failures}
\end{figure}

\section{Discussion}
We present the Panoptic Studio and an interaction capture method that leverages a large number of views. To demonstrate the performance of our method, we collect a large scale social interaction dataset, and produce compelling motion capture results on it. In particular, we empirically find that having a larger number of views is more beneficial than having a higher resolution of views for social interaction capture. Our quantitative comparison on various number and type of cameras can be used as a meaningful resource to design follow-up multiview systems to estimate the required number of views to achieve a desired accuracy. Our method also demonstrates that highly-occluded social motion capture is possible by boosting 2D pose detection cues and motion cues in a larger number of views, without using any heavy prior or template model. Our method shows its advantages in the social interaction capture scenario by reconstructing subjects of diverse appearance, body sizes, and body topology for a long term without error accumulation issue.

\subsection{Limitations}
\label{subsection:Limitation}
A limitation of our method is the dependency on a 2D pose detection method. State-of-the-art pose detectors are weak in detecting unusual poses and closely located people (as shown in  Fig.~\ref{fig:failures}). We also find that the pose detector sometimes gets confused in distinguishing left-right limbs (as shown in the center of Figure~\ref{fig:failures}). Although our method can overcome these issues by fusing cues across views via spatial voting and across time via associating trajectory streams, if the 2D body pose detectors fail consistently, our method is unable to recover. The second limitation is the long computation time to process the large number of views. However, depending on the application, the computation time can be greatly reduced by using a smaller number of cameras with a trade-off in accuracy (see Fig.~\ref{fig:quant1}). 

In terms of the hardware design, we also find two limitations which should be reconsidered for follow-up research. The first is the incompatible frame rates among heterogeneous sensors, especially between HD cameras and VGA cameras, which makes it hard to fuse them for 3D reconstruction. Due to this reason, our method currently uses VGA cameras only, but we expect that this issue can be dealt with by interpolating cues in  the common time domain, because all the sensors are temporally aligned in millisecond level. The second issue is that all the camera views mainly focus on the center of the dome and, thus, fewer views are available at the edges of the capture volume. Such design is ideal given the assumption that subjects are located at the center of the system, but we observe that sometimes people tend to stand near the walls during social interactions. An alternative direction would be to make cameras focus on random locations so that view coverage can be uniformly spread throughout the working volume. 

\subsection{Future Work}
There are various future directions expanding our system and outputs. First, analyzing human social behaviors using the measured social signals of our system is an interesting direction, which will facilitate social behavior understanding in a data-driven manner. Second, our interaction capture outputs can be used as labeled data to train new 2D detectors. By projecting the skeletal reconstruction outputs on all 521 views, our current dataset generates 153 million pose data in diverse views. Although the appearance diversity of the scenes would be limited compared to internet photos, our dataset is still meaningful in that: (1) it captures multiple interacting people showing severe inter-occlusions in the scenes; (2) it has annotations for entire video frames which is the key to study temporal relation of poses; (3) the scenes are taken by 521 diverse view points compared to the biased views of usual portrait photos (e.g., frontal views or side views). This type of labeled data would be hard to obtain by manual annotations. Lastly, a similar massively multiview approach can be applied to reconstruct 3D faces. This can be done by substituting the 2D body pose detector with a 2D face landmark detector in our method.

\ifCLASSOPTIONcaptionsoff
  \newpage
\fi



%
\bibliographystyle{IEEEtran}
\bibliography{egbib}

\end{document}